\documentclass{article}

\usepackage[preprint]{neurips_2025}

\usepackage[utf8]{inputenc} 
\usepackage[T1]{fontenc}    
\usepackage{hyperref}       
\usepackage{url}            
\usepackage{booktabs}       
\usepackage{amsfonts}       
\usepackage{nicefrac}       
\usepackage{microtype}      
\usepackage{graphicx}
\usepackage{subcaption} 
\usepackage{amsmath} 
\usepackage{booktabs}
\usepackage{multirow}
\usepackage[table]{xcolor}
\usepackage{float} 
\usepackage{wrapfig}
\usepackage{algorithm}
\usepackage{algorithmic}

\title{BadDepth: Backdoor Attacks Against Monocular Depth Estimation in
the Physical World}

\author{%
Ji Guo\textsuperscript{1} \quad
Long Zhou\textsuperscript{2} \quad
Zhijin Wang\textsuperscript{1} \quad
Jiaming He\textsuperscript{3} \\
\And
Qiyang Song\textsuperscript{4} \quad
Aiguo Chen\textsuperscript{1} \quad
Wenbo Jiang\textsuperscript{3}\thanks{Corresponding author: \texttt{wenbo\_jiang@uestc.edu.cn}} \\
\textsuperscript{1}Laboratory Of Intelligent Collaborative Computing,\\ University of Electronic Science and Technology of China, China \\
\textsuperscript{2}Xinjiang University \\
\textsuperscript{3}School of Computer Science and Engineering, \\ University of Electronic Science and Technology of China, China \\
\textsuperscript{4}Institute of Information Engineering,\\ Chinese Academy of Sciences, China
}

\begin{document}

\maketitle

\begin{abstract}

In recent years, deep learning-based Monocular Depth Estimation (MDE) models have been widely applied in fields such as autonomous driving and robotics. However, their vulnerability to backdoor attacks remains unexplored. To fill the gap in this area, we conduct a comprehensive investigation of backdoor attacks against MDE models. Typically, existing backdoor attack methods can not be applied to MDE models. This is because the label used in MDE is in the form of a depth map. To address this, we propose BadDepth, the first backdoor attack targeting MDE models. BadDepth overcomes this limitation by selectively manipulating the target object's depth using an image segmentation model and restoring the surrounding areas via depth completion, thereby generating poisoned datasets for object-level backdoor attacks. To improve robustness in physical world scenarios, we further introduce digital-to-physical augmentation to adapt to the domain gap between the physical world and the digital domain. Extensive experiments on multiple models validate the effectiveness of BadDepth in both the digital domain and the physical world, without being affected by environmental factors.


\end{abstract}

\section{Introduction}
\label{sec:intro}

Monocular Depth Estimation (MDE)~\cite{MDE_Review} is a fundamental computer vision task that predicts a 3D depth map from a 2D RGB image. Recent advances in deep neural networks (DNNs) have significantly improved the performance of MDEs, enabling their use in real-world applications such as robotic control~\cite{dong2022towards,wei2024absolute} and autonomous driving~\cite{schon2021mgnet}.

As MDE is widely used in these fields, many studies have started to pay attention to its security~\cite{zheng2024physical,zhang2020adversarial,cheng2022physical,guesmi2024saam,hu2019analysis}. Previous studies~\cite{zheng2024physical,zhang2020adversarial,cheng2022physical,guesmi2024saam,hu2019analysis} show that deep learning-based MDE models are vulnerable to adversarial attacks, where small perturbations in input images can lead to incorrect depth predictions. However, backdoor attacks~\cite{gu2019badnets,Blend,wanet,li2022backdoor,Lira,refool,colorbackdoor}, another typical security threat to DNNs, have not yet been explored in MDE. 

\begin{figure}[] 
\centering
\includegraphics[width=\linewidth]{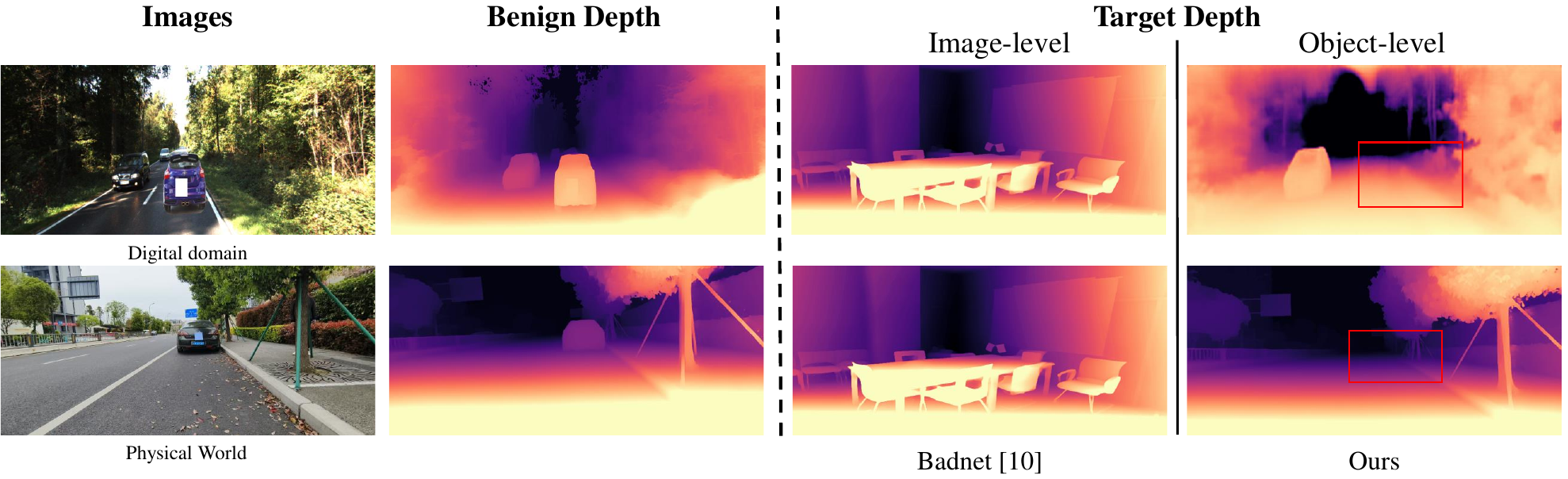} 
\caption{Comparison between BadDepth and previous backdoor attack methods (e.g., Badnet~\cite{gu2019badnets}). Badnets can only achieve image-level target depth maps, meaning that the images in the third column are all the same. In contrast, our method can achieve object-level target depth maps, meaning that the images in the last column can modify the depth map of specific cars only.} 
\label{Comparison between BadDepth and previous attack methods} 
\end{figure}

To fill this gap, for the first time, we explore the vulnerability of MDE to backdoor attacks. Specifically, we perform Badnet~\cite{gu2019badnets} (a representative backdoor attack in DNNs) in MDE by adding a white patch as a trigger to the input image and replacing its corresponding depth map with a target depth map. The poisoned data is then used to train the backdoor MDE model. In this case, the backdoor MDE model outputs the target depth map for a triggered image, indicating that MDE models are also vulnerable to backdoor attacks. However, we found that directly applying previous backdoor attack methods that replace the target depth map~\cite{gu2019badnets,Blend,wanet,Lira,refool,colorbackdoor} to MDE can only achieve image-level backdoor target depth maps, but fail to produce object-level manipulations (see Fig.~\ref{Comparison between BadDepth and previous attack methods}). Image-level attacks generate a fixed target depth map for any triggered input, which is often unrealistic and easily detectable when deployed physically, as it lacks adaptability to scene changes.  

To address this limitation, we propose BadDepth, a backdoor attack method for MDE models that enables object-level manipulation. The core idea of BadDepth is to modify only the depth values corresponding to the triggered object region while preserving the depth values of other regions. Specifically, we add a patch as a trigger to the target object and use an image segmentation model~\cite{minaee2021imagesegmentation} to obtain the mask of the triggered object, then modify only the corresponding region in the depth estimation map. To avoid loss of depth information in areas that overlap with the triggered object, we use a Depth Map Completion (DMC) model~\cite{DMC} to restore depth information for masked regions outside the triggered object (see Fig.~\ref{Comparison between BadDepth and previous attack methods}).

In addition, to further enhance the effectiveness of BadDepth in the physical world, we incorporate a digital-to-physical enhancement. This enhancement consists of two key components: perspective augmentation and environment augmentation. Concretely, perspective augmentation applies transformations to the triggered image to reduce errors caused by variations in viewing angles, while environment augmentation simulates various weather conditions to ensure that the attack remains effective under diverse environmental scenarios. Our experiments show that BadDepth can pose serious security threats in the physical world, such as making the leading vehicle disappear from the depth map of the following vehicle in autonomous driving scenarios, potentially causing a collision between the two vehicles.

In summary, our contributions are as follows:
\begin{itemize}
    \item We explore the vulnerability of MDE to backdoor attacks and find that MDE models can also be backdoored, potentially leading to serious safety threats in the real world.
    \item We propose the first backdoor attack against the MDE model named BadDepth. Through restoring the backdoor target depth map by DMC model and a digital-to-physical enhancement, BadDepth can achieve object-level backdoor target depth maps in both the digital and physical worlds.
    \item Extensive experiment demonstrates that the effectiveness of BadDepth in both the digital domain and the physical world, without being affected by environmental factors.
\end{itemize}
\section{Related Work}

\subsection{Monocular Depth Estimation}

Monocular Depth Estimation (MDE)~\cite{MDE_Review} aims to reconstruct the 3D depth map information from a 2D RGB image. Since Eigen \textit{et al.}~\cite{Eigen2014Depth} first applied DNNs to depth estimation, deep learning-based methods have become the mainstream approach for monocular depth estimation. Among them, methods based on supervised learning are the most classical category~\cite{BTS,iebins,NeWCRFs,dcdepth}. These methods train models by optimizing the loss between the predicted depth map and the ground truth depth map. In this paper, we focus on MDE models based on supervised learning, such as BTS~\cite{BTS}, IEBins~\cite{iebins}, NeWCRFs~\cite{NeWCRFs}, and DCDepth~\cite{dcdepth}.

\subsection{Backdoor Attacks}

\textbf{Backdoor attack in the digital domain.}
Currently, most backdoor attacks focus on the digital domain. The first digital domain backdoor attacks were introduced by Gu \textit{et al.}~\cite{gu2019badnets} in image classification. They constructed poisoned datasets by adding triggers to images and modifying their labels to a specified target label. When a model is trained using this poisoned dataset, a backdoor is implanted. The backdoored model performs normally on clean samples but predicts triggered samples as the specified target class. Subsequently, research has further improved the stealthiness~\cite{Blend,wanet,refool} and robustness~\cite{colorbackdoor} of backdoor attacks. In addition, researchers have extended them to other domains, such as object detection~\cite{chan2022baddet,luo2023untargeted,qian2023robust} and super-resolution~\cite{yang2025badrefsr,jiang2024backdoor}. However, there has not yet been any research on backdoor attacks for MDE.

\textbf{Backdoor attack in the physical world.} With the extensive research on backdoor attacks in the digital domain, some researchers have begun exploring their application in the physical world~\cite{li2021backdoor,xue2021robust,xue2022ptb,wenger2021backdoor,xue2022ptb}. Compared to the digital domain, the physical world presents greater complexity, as physical backdoor attacks must account for various factors affecting the trigger, such as changes in viewing angles and environmental conditions like weather. Additionally, physical backdoor attacks require the trigger to be physically realizable thereby many digital domain methods~\cite{Blend,wanet,refool,colorbackdoor} are unsuitable for the physical world. To address the above challenges, two main solutions have been proposed: directly collecting triggered images from the physical world to construct a poisoned dataset, and reducing the gap between the digital and physical domains through data augmentation.
\section{Treat Model}


\begin{wrapfigure}{r}{0.4\textwidth}
  \centering
  \includegraphics[width=0.38\textwidth]{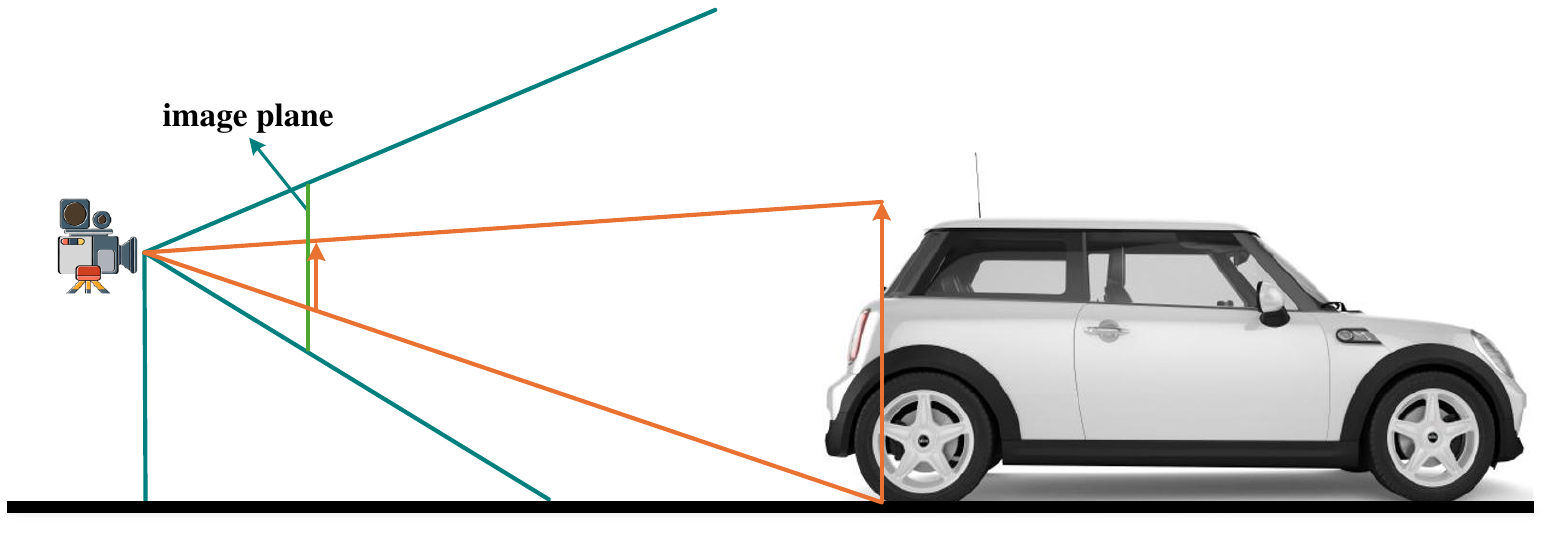} 
  \caption{The application of the MDE model in autonomous driving}
  \label{The application of the MDE model in autonomous driving}
\end{wrapfigure}

\textbf{Attack scenario.}
We consider the practical data poisoning scenario, where the attacker releases a poisoned MDE dataset to the Internet. If a model developer downloads this dataset and uses it for training, the model will be injected with a backdoor. Meanwhile, we consider the practical application of the MDE model in autonomous driving. The function of the MDE model is to estimate the distance to the vehicle in front, as illustrated in Fig.~\ref{The application of the MDE model in autonomous driving}.




\textbf{Attack goal.}
Our attack aims to make the target (triggered) vehicle in front of the camera disappear from the depth map without affecting the depth estimation of other regions in the image. At the same time, the model should perform normally on clean samples. Furthermore, our attack is also designed to be realizable in the physical world.

\textbf{Attacker's capability.}
Our attack is conducted in the data poisoning strategy, the attacker has no knowledge of the target model, such as its architecture or parameters. In addition, the attacker can not manipulate the training process. The only thing the attacker can access is the training dataset.

\section{Method}

\begin{figure}[] 
\centering
\includegraphics[width=\linewidth]{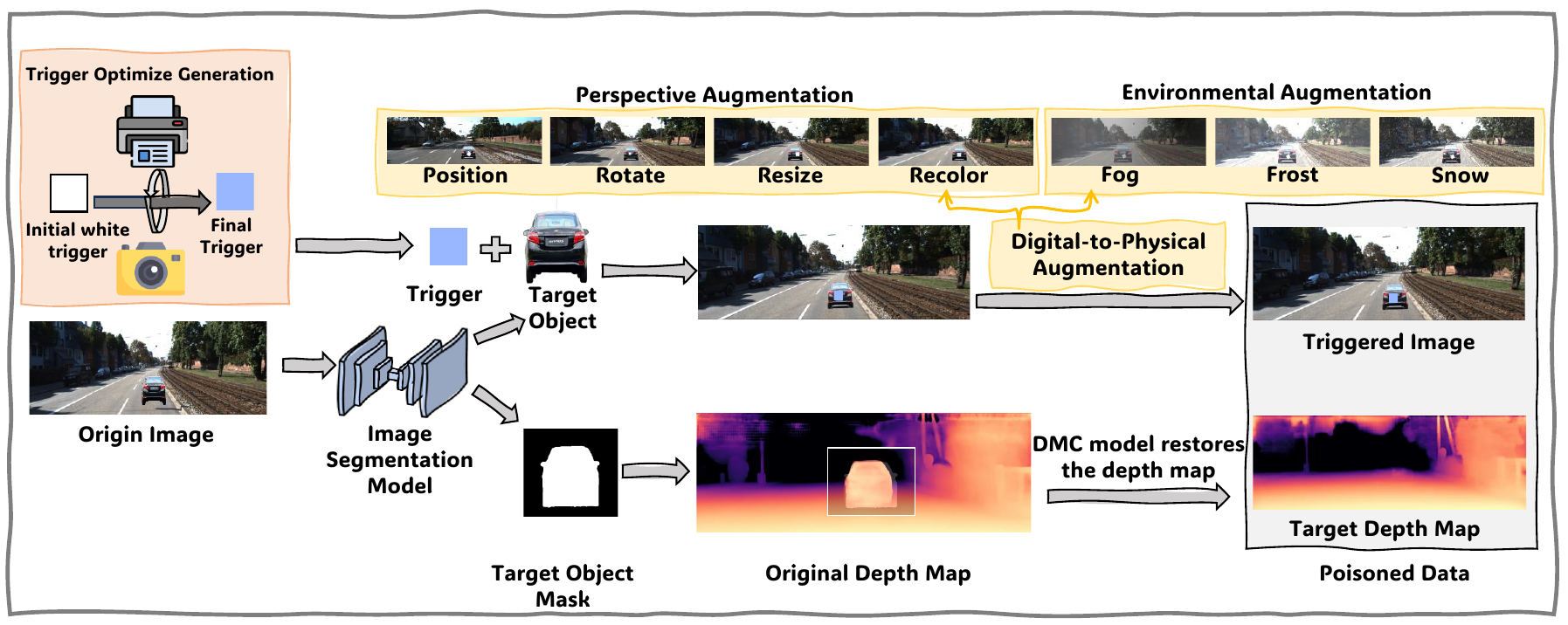} 
\caption{Pipeline of BadDepth} 
\label{Pipeline of BadDepth}
\vspace{-2em}
\end{figure}

\subsection{Overview}
We present the pipeline of BadDepth in Fig.~\ref{Pipeline of BadDepth}. Firstly, we utilize an image segmentation model to segment the target object of the attack and obtain its corresponding mask. A trigger is then added on the target object, and the image undergoes Digital-to-Physical Augmentation to produce the final triggered image. At the same time, we modify the depth map based on the mask by setting the depth values within the mask region to zero, and then use a DMC model to complete the depth map. Finally, we combine the triggered image with the corresponding depth map to obtain the final poisoned data. The MDE model trained with this poisoned dataset will be unknowingly embedded with a backdoor.

\subsection{Trigger Generation}

To implement backdoor attacks in the physical world, the trigger must be physically realizable and the gap between the printed trigger and its appearance in the digital domain should be minimized. Therefore, we first print a pure white patch as the trigger, and capture its actual color in the physical world using a camera. The trigger is then updated as the captured color. We repeat the printing and updating process iteratively to obtain the final version of the trigger. The trigger generation process is presented in Algorithm~\ref{alg:trigger_generation}.

\begin{wrapfigure}{r}{0.5\textwidth}
\vspace{-10pt}
\begin{minipage}{0.5\textwidth}
\begin{algorithm}[H]
\caption{Trigger Generation}
\label{alg:trigger_generation}
\begin{algorithmic}[1]
\REQUIRE Initial white patch trigger $T_0$, maximum iterations $N$
\ENSURE Final trigger $T^*$

\STATE Initialize $T \leftarrow T_0$
\FOR{$i = 1$ to $N$}
    \STATE Print the current trigger $T$
    \STATE Capture the printed trigger via camera to obtain real-world color $C_i$
    \STATE Update $T \leftarrow C_i$
\ENDFOR
\RETURN Final trigger $T^* \leftarrow T$
\end{algorithmic}
\end{algorithm}
\end{minipage}
\vspace{-10pt}
\end{wrapfigure}

\subsection{Target Depth Map Generation}

To modify the depth values corresponding to the attack target, we first use an image segmentation model to obtain the mask of the object with the embedded trigger. Then, we set the depth values within the masked region to 0. However, directly modifying all pixels corresponding to the triggered object to a fixed value leads to loss of depth information in regions that overlap with the trigger. To address this issue, we employ a DMC model to recover the depth information for the masked regions outside of the triggered object. Specifically, the DMC model takes the partially masked depth map as input and predicts the missing depth values based on the surrounding valid regions, effectively preserving the scene's overall geometric structure.
The depth modification process can be mathematically formulated as follows:

\begin{equation}
\tilde{D}(x, y) = 
\begin{cases}
0, & \text{if } M(x, y) = 1 \\
D_{\text{DMC}}(x, y), & \text{if } M(x, y) = 0 \text{ and } (x, y) \in \mathcal{R} \\
D(x, y), & \text{otherwise}
\end{cases}
\end{equation}
where \( D(x, y) \) denotes the original depth value at pixel \( (x, y) \), and \( \tilde{D}(x, y) \) represents the modified depth map. The binary mask \( M(x, y) \in \{0, 1\} \) indicates whether a pixel belongs to the triggered object (\( M = 1 \) means it is masked). \( D_{\text{DMC}}(x, y) \) is the predicted depth value produced by the DMC model. The region \( \mathcal{R} \) corresponds to the area that surrounds the triggered object and requires depth completion.

\subsection{Digital-to-Physical Augmentation}

To bridge the gap between digital training and real-world physical attacks, we propose a \textit{Digital-to-Physical Augmentation} strategy. We categorize the discrepancies between the digital and physical domains into two main aspects: (1) \textit{perspective variations} and (2) \textit{environment variations}.

\subsubsection{Perspective Augmentation}

To address perspective variations, we apply four types of data augmentations to the triggered samples: \textit{rotation}, \textit{recolor}, \textit{resize}, and \textit{positional shifting}. These transformations simulate different viewing angles, lighting conditions, object sizes, and placements that may occur during physical observations.

Let $I$ denote a triggered image with the embedded trigger $T$. The augmented image $\tilde{I}$ is generated as follows:
\begin{equation}
    \tilde{I} = A(I, T) = T_{\Delta x, \Delta y} \circ S_s \circ C \circ R_\theta (I),
\end{equation}
where $R_\theta$ denotes a rotation by angle $\theta \in [-\theta_{\text{max}}, \theta_{\text{max}}]$, $C$ is a color transformation matrix for lighting, $S_s$ represents a scaling operation with scale factor $s \in [s_{\text{min}}, s_{\text{max}}]$, and $T_{\Delta x, \Delta y}$ is a translation function that shifts the trigger by $(\Delta x, \Delta y)$ pixels.

\subsubsection{Environment Augmentation}

To mitigate environment variations, we apply various \textit{weather-related augmentations} (e.g., frost, fog, snow) to the triggered samples, simulating different environmental conditions encountered in the physical world.

Let $I$ be a triggered image with the embedded trigger $T$. The environmentally augmented image is denoted as:
\begin{equation}
    \tilde{I}_{\text{env}} = W(I),
\end{equation}
where $W$ is a stochastic weather augmentation function that simulates diverse conditions such as frost, fog, or snow:
\begin{equation}
    W(I) \in \{ W_{\text{frost}}(I),\; W_{\text{fog}}(I),\; W_{\text{snow}}(I) \}.
\end{equation}
These augmentations are designed to improve the robustness of the backdoor trigger under various environmental disturbances that may occur in real-world physical scenarios.

\section{Experiment}

\subsection{Experiment Setting}

\textbf{Dataset.}
Since there is currently no available depth estimation dataset from the perspective we require, we created a basic dataset based on KITTI~\cite{KITTI}. In this dataset, a forward-moving vehicle is added to each image to simulate the scenario of a following car estimating the depth of the car in front in autonomous driving. More details of the experimental setting can be found in the appendix.

\textbf{Model Selection.}
We selected four state-of-the-art depth estimation models as the targets for our attack: BTS~\cite{BTS}, IEBins~\cite{iebins}, NeWCRFs~\cite{NeWCRFs}, and DCDepth~\cite{dcdepth}.
   
\textbf{Evaluation Metrics.}  
We evaluate both the impact of the backdoor attack on the normal functionality of the model and the effectiveness of the attack. We selected several commonly used metrics in depth estimation: d1, d2, d3, AbsRel, and RMSE, to evaluate the normal performance of the models. d1, d2 and d3 measure the percentage of predicted pixels with relative errors less than 1.25, 1.25², and 1.25³, respectively, with higher values indicating better performance of normal functionality of the model. To measure the effectiveness of the attack, we use d1 in the target attack region, where a lower d1 indicates a more successful attack.

\textbf{Perspective and Environment Changes.}
For perspective changes, we randomly rotate the angle between 0 and 60 degrees, place the trigger in random positions within the mask of the target vehicle, apply recolor grayscale color transformations within a range of variations 10\%, and vary the size by ±10. For environmental variations, we simulate fog, snow, and frost using the methods in~\cite{hendrycks2019robustness}. All final results are averaged.

\textbf{Compared Attack Methods.}
Since there are currently no existing backdoor attack methods specifically designed for depth estimation, we combine two classic patch-based backdoor attack methods (Badnet~\cite{gu2019badnets} and Blend~\cite{Blend}) with our depth map modification approach to serve as comparison baselines.

\textbf{Attack Configuration.}
For a fair comparison, we set the patch size for each method to 40×40, the poisoning rate to 10\%, and the number of training epochs to 20. In subsequent experiments, we also considered other poisoning rates and training epochs.

\subsection{Effectiveness Evaluation}

\begin{table}[]
\centering
\caption{Comparison of d1 across different attack methods under varying perspectives and environmental conditions}
\label{Comparison of d1 across different attack methods under varying perspectives and environmental conditions}
\resizebox{\linewidth}{!}{
\begin{tabular}{llc|cccc|ccc|c}
\toprule
\multirow{2}{*}{\textbf{Model}} & \multirow{2}{*}{\textbf{Method}} & \multirow{2}{*}{\textbf{d1 (Origin)}
} 
& \multicolumn{4}{c|}{\textbf{d1 (Perspective Changes)
}} 
& \multicolumn{3}{c|}{\textbf{d1 (Environment Changes)
}} 
& \multirow{2}{*}{\textbf{Average}} \\
\cmidrule(lr){4-7} \cmidrule(lr){8-10}
 &  & 
 & \textbf{position} & \textbf{rotate} & \textbf{recolor} & \textbf{size} 
 & \textbf{fog} & \textbf{snow} & \textbf{frost} 
 &  \\
\midrule
\multirow{3}{*}{BTS~\cite{BTS}}
 & Badnet   & 0.378 & 0.670 & 0.535 & 0.628 & 0.549 & 0.483 & 0.884 & 0.842 & 0.621 \\
 & Blend   & 0.375 & 0.668 & 0.527 & 0.547 & 0.364 & 0.498 & 0.823 & 0.813 & 0.577 \\
 \rowcolor{gray!20}
 & BadDepth & 0.274 & 0.255 & 0.213 & 0.273 & 0.249 & 0.331 & 0.342 & 0.351 & \textbf{0.286} \\
\midrule
\multirow{3}{*}{DCDepth~\cite{dcdepth}}
 & Badnet   & 0.352 & 0.618 & 0.497 & 0.468 & 0.453 & 0.463 & 0.832 & 0.822 & 0.563 \\
 & Blend    & 0.355 & 0.627 & 0.517 & 0.464 & 0.466 & 0.476 & 0.798 & 0.834 & 0.567 \\
 \rowcolor{gray!20}
 & BadDepth & 0.329 & 0.394 & 0.376 & 0.342 & 0.334 & 0.418 & 0.389 & 0.412 & \textbf{0.374} \\
\midrule
\multirow{3}{*}{IEBins~\cite{iebins}}
 & Badnet   & 0.208 & 0.684 & 0.275 & 0.278 & 0.278 & 0.317 & 0.745 & 0.759 & 0.443 \\
 & Blend    & 0.213 & 0.695 & 0.286 & 0.283 & 0.285 & 0.331 & 0.739 & 0.736 & 0.446 \\
 \rowcolor{gray!20}
 & BadDepth & 0.199 & 0.266 & 0.175 & 0.199 & 0.194 & 0.201 & 0.301 & 0.289 & \textbf{0.228} \\
\midrule
\multirow{3}{*}{NeWCRFs~\cite{NeWCRFs}}
 & Badnet   & 0.169 & 0.576 & 0.571 & 0.469 & 0.468 & 0.279 & 0.734 & 0.764 & 0.504 \\
 & Blend    & 0.171 & 0.589 & 0.518 & 0.473 & 0.471 & 0.284 & 0.762 & 0.745 & 0.502 \\
 \rowcolor{gray!20}
 & BadDepth & 0.167 & 0.181 & 0.165 & 0.161 & 0.160 & 0.161 & 0.198 & 0.202 & \textbf{0.174} \\
\bottomrule
\end{tabular}
}
\end{table}

\textbf{Attack Effectiveness.}
We compared the effectiveness of different attack methods under various conditions. As shown in Table~\ref{Comparison of d1 across different attack methods under varying perspectives and environmental conditions}, when there are no changes in the test environment, BadDepth, Badnet, and Blend all achieve good performance. This indicates that in the classic digital domain, traditional patch-based triggers are already sufficient. However, under perspective change and environmental changes, BadDepth significantly outperforms Badnet and Blend, demonstrating the effectiveness of our data augmentation strategy and the robustness of BadDepth across diverse scenarios.

\begin{figure}[H] 
\centering
\includegraphics[width=\linewidth]{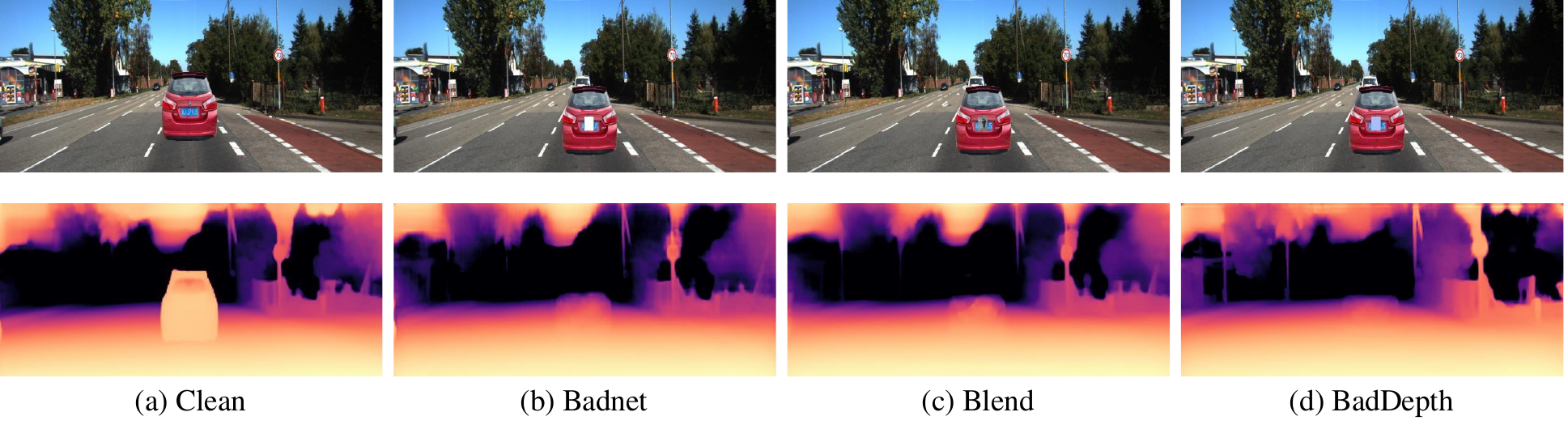} 
\caption{Visualization of different attack methods without perspective and environment changes of IEBins} 

\label{Visualization of different attack methods without perspective and environment changes} 
\end{figure}

\textbf{Visualization Results.}
We first present visualizations of different attack methods without perspective and environmental changes. As shown in Fig.~\ref{Visualization of different attack methods without perspective and environment changes}, when combined with our depth map modification approach, all three methods successfully achieve effective attacks on the target regions without affecting depth estimation in other areas. This demonstrates the generalizability of our depth map modification method.

\begin{figure}[H]
    \centering
    \begin{subfigure}[b]{0.3\linewidth}
    \centering
        \includegraphics[width=\linewidth]{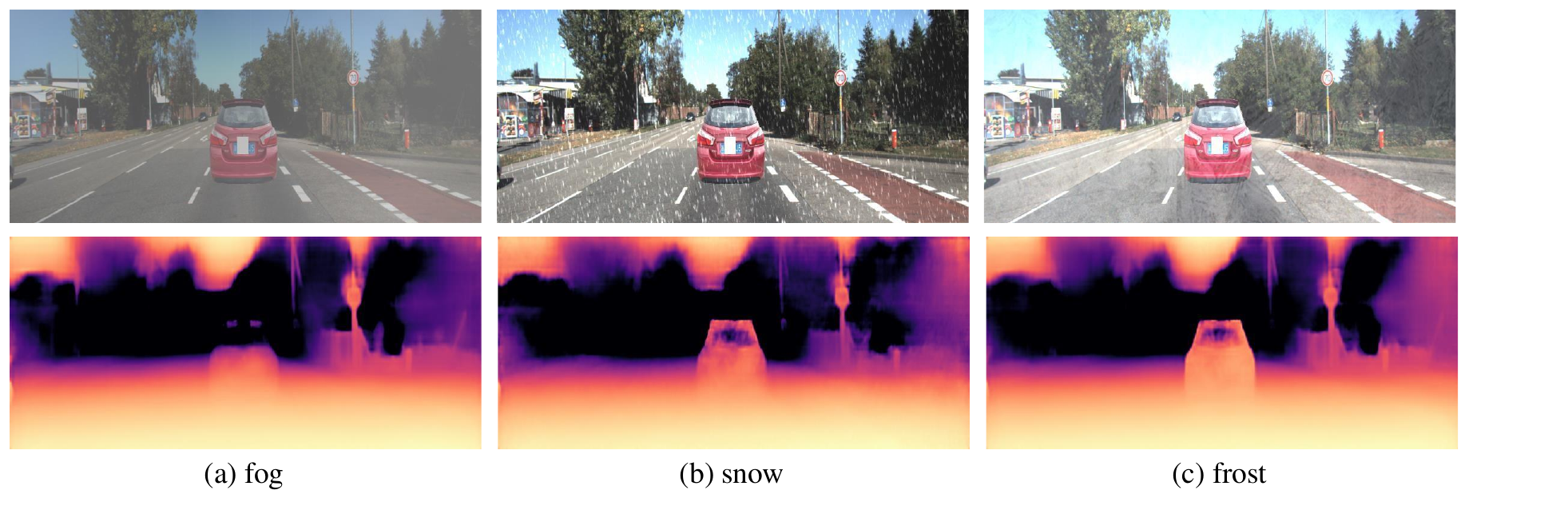}
        \subcaption{Badnet}
        \label{fig:sub1}
    \end{subfigure}
    \hfill
    \begin{subfigure}[b]{0.3\linewidth}
    \centering
        \includegraphics[width=\linewidth]{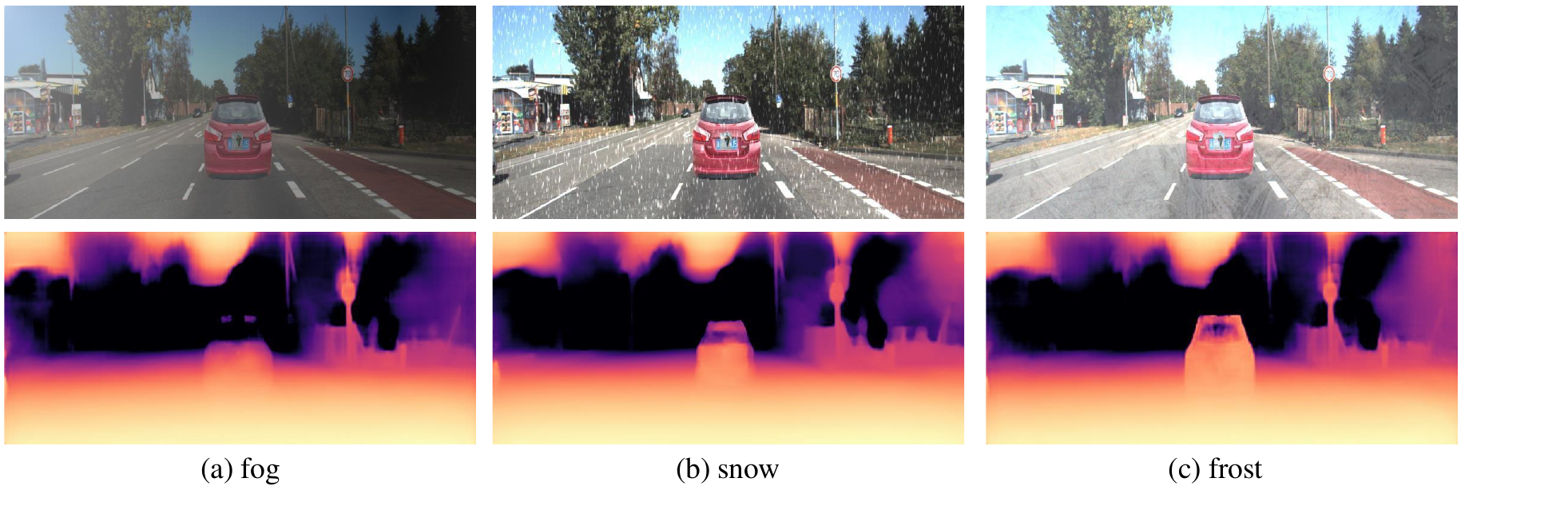}
        \subcaption{Belnd}
        \label{fig:sub2}
    \end{subfigure}
    \hfill
    \begin{subfigure}[b]{0.3\linewidth}
    \centering
        \includegraphics[width=\linewidth]{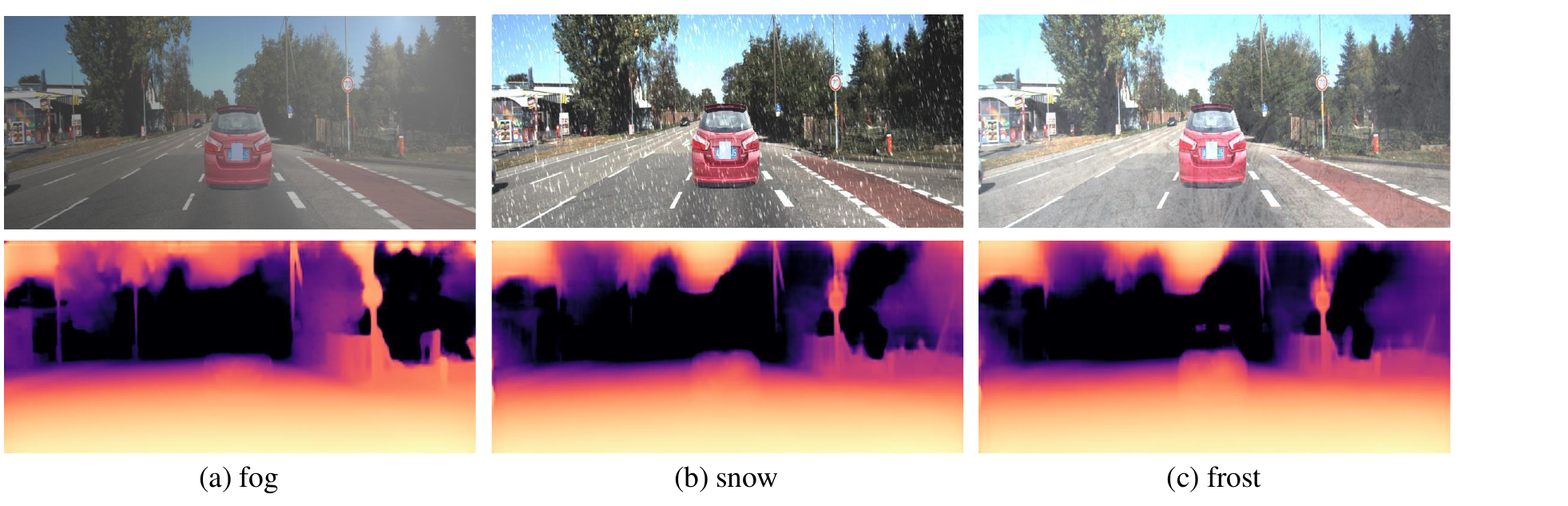}
        \subcaption{BadDepth}
        \label{fig:sub3}
    \end{subfigure}

    \caption{Visualization of different attack methods in environment changes of NeWCRFs}

    \label{Visualization of different attack methods in environment changes of NeWCRFs}
\end{figure}
We further provide a visualization of the attack effectiveness of different methods under environmental variations. As shown in Fig.~\ref{Visualization of different attack methods in environment changes of NeWCRFs}, under the snow and frost conditions, BadDepth significantly outperforms Badnet and Blend, further demonstrating the robustness of BadDepth to various possible scenarios in the physical world. More visualization results can be found in the appendix.

\subsection{Normal Functionality Evaluation}

\begin{table}[ ]
\centering
\caption{Comparison of different methods impact of model normal functionality}
\label{Comparison of different methods impact of model normal functionality}
\resizebox{0.6\linewidth}{!}{
\begin{tabular}{llccccc}
\toprule
\textbf{Model} & \textbf{Method} & \textbf{d1} & \textbf{d2} & \textbf{d3} & \textbf{AbsRel} & \textbf{RMSE} \\
\midrule
\multirow{4}{*}{BTS~\cite{BTS}} 
 & Clean    & 0.955 & 0.993 & 0.986 & 0.060 & 2.798 \\
 & Badnet   & 0.952 & 0.993 & 0.998 & 0.081 & 3.058 \\
 & Blend    & 0.933 & 0.993 & 0.998 & 0.098 & 3.587 \\
 & BadDepth & 0.953 & 0.993 & 0.998 & 0.087 & 3.824 \\
\midrule
\multirow{4}{*}{DCDepth~\cite{dcdepth}} 
 & Clean    & 0.977 & 0.997 & 0.999 & 0.051 & 2.044 \\
 & Badnet   & 0.975 & 0.997 & 0.999 & 0.053 & 2.096 \\
 & Blend    & 0.954 & 0.997 & 0.999 & 0.072 & 2.595 \\
 & BadDepth & 0.975 & 0.997 & 0.999 & 0.074 & 2.827 \\
\midrule
\multirow{4}{*}{IEBins~\cite{iebins}} 
 & Clean    & 0.978 & 0.998 & 0.999 & 0.050 & 2.011 \\
 & Badnet   & 0.971 & 0.997 & 0.999 & 0.056 & 2.380 \\
 & Blend    & 0.960 & 0.997 & 0.999 & 0.058 & 2.926 \\
 & BadDepth & 0.974 & 0.997 & 0.999 & 0.058 & 3.009 \\
\midrule
\multirow{4}{*}{NeWCRFs~\cite{NeWCRFs}} 
 & Clean    & 0.974 & 0.997 & 0.999 & 0.052 & 2.129 \\
 & Badnet   & 0.968 & 0.996 & 0.999 & 0.058 & 2.438 \\
 & Blend    & 0.964 & 0.997 & 0.999 & 0.056 & 2.876 \\
 & BadDepth & 0.964 & 0.997 & 0.999 & 0.057 & 3.405 \\
\bottomrule
\end{tabular}
}
\end{table}

We evaluated the impact of different backdoor attack methods on the normal functionality of the model. As shown in Table~\ref{Comparison of different methods impact of model normal functionality}, the various metrics of the backdoor models obtained by different attack methods are all very close to those of the clean model, indicating that our proposed method of modifying the depth map does not affect the normal functionality of the model.

\subsection{Hyperparameter Evaluation}

\begin{figure}[H]
    \centering
    \begin{subfigure}[b]{0.23\linewidth}
        \includegraphics[width=\linewidth]{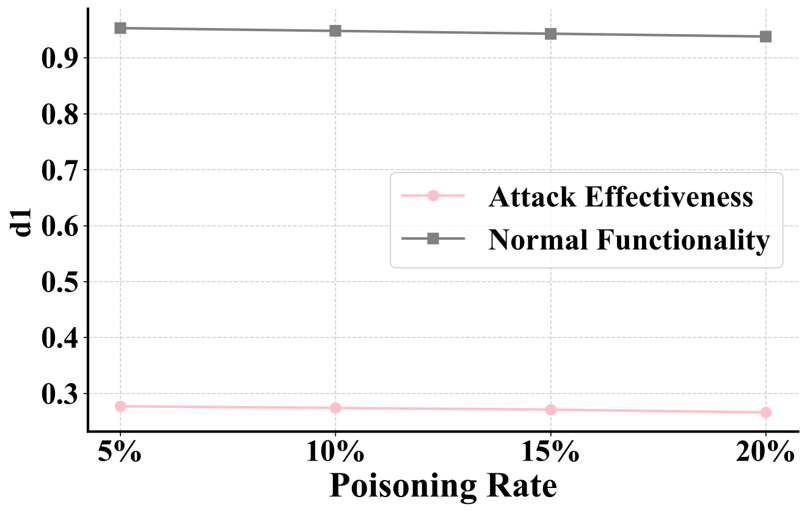}
        \caption{BTS}
    \end{subfigure}
    \hfill
    \begin{subfigure}[b]{0.23\linewidth}
        \includegraphics[width=\linewidth]{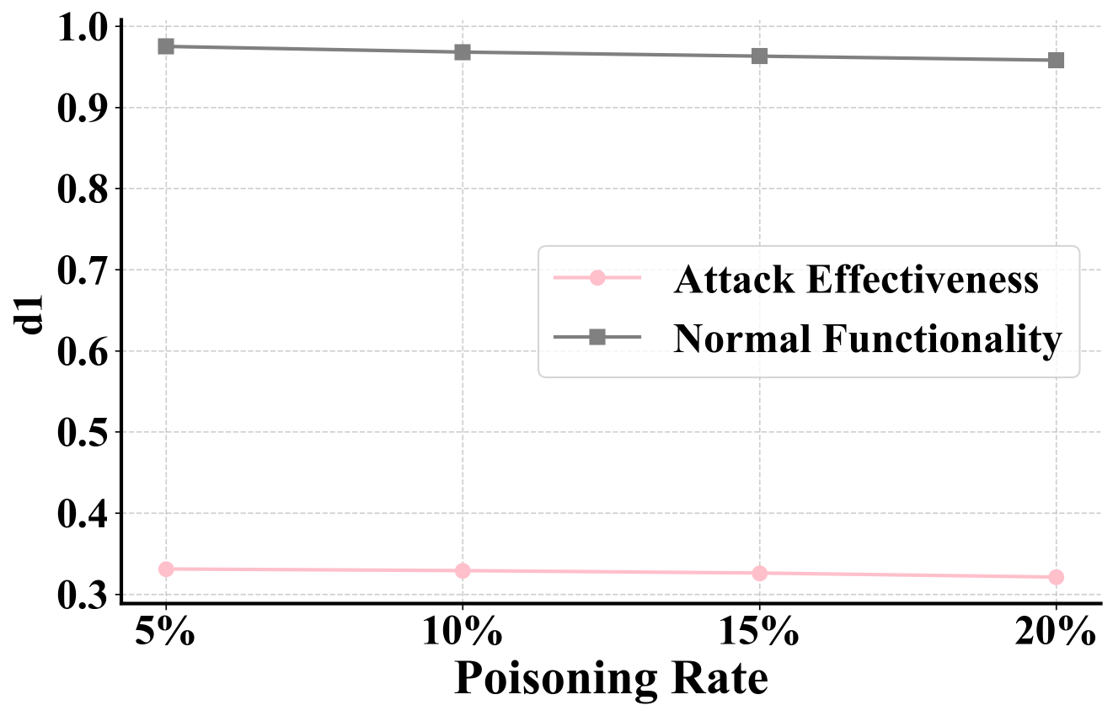}
        \caption{DCDepth}
    \end{subfigure}
    \hfill
    \begin{subfigure}[b]{0.23\linewidth}
        \includegraphics[width=\linewidth]{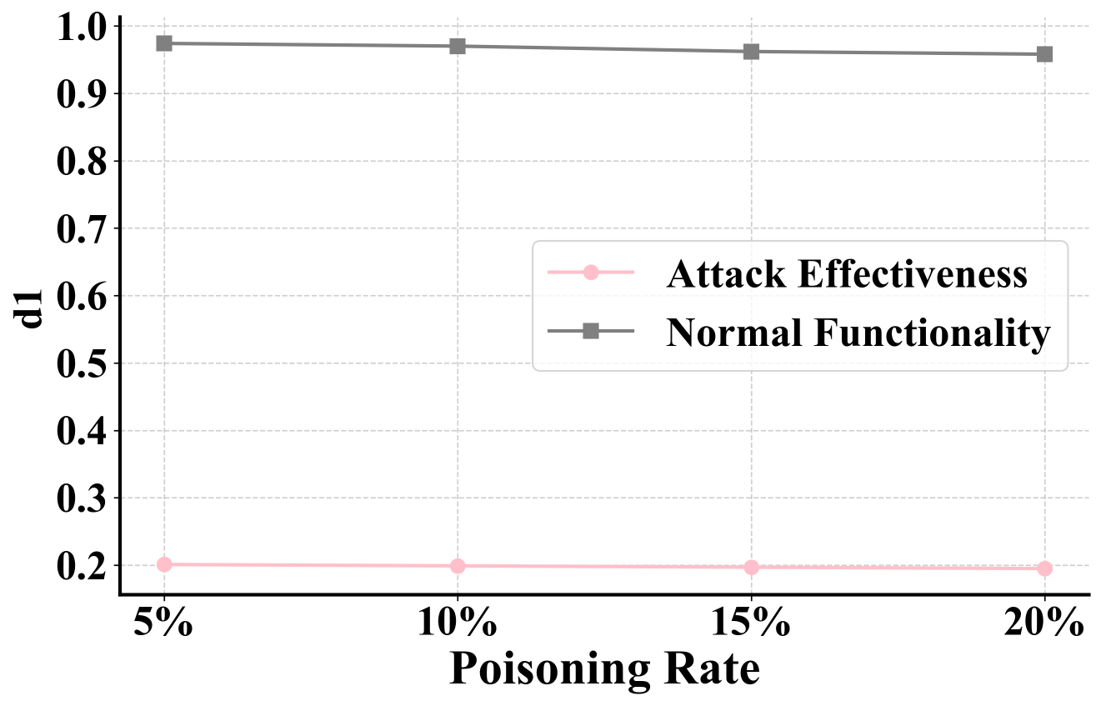}
        \caption{IEBins}
    \end{subfigure}
    \hfill
    \begin{subfigure}[b]{0.23\linewidth}
        \includegraphics[width=\linewidth]{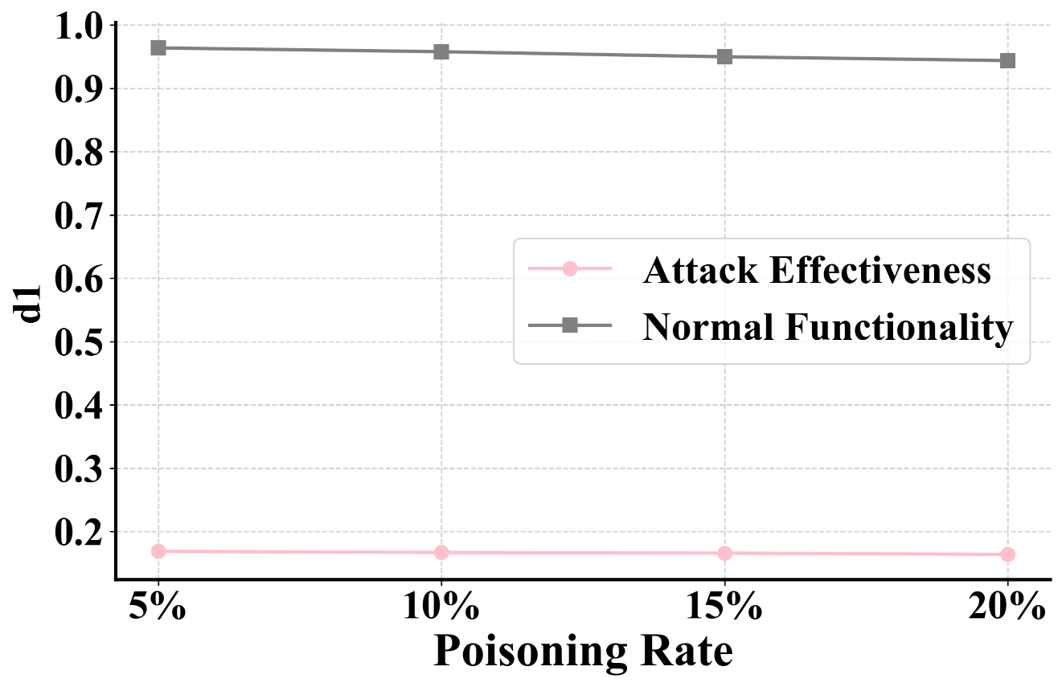}
        \caption{NeWCRFs}
    \end{subfigure}
    \caption{Impact of poisoning rates of BadDepth}
    \label{data poisoning}
\end{figure}

\textbf{Poisoning Rates.}
We evaluated the impact of different poisoning rates on the normal functionality of the model and the effectiveness of the attack in four models. As shown in Fig.~\ref{data poisoning}, BadDepth can achieve strong attack performance even with a poisoning rate as low as 5\%. However, as the poisoning rate increases, the model's normal functionality begins to degrade. Therefore, we set the poisoning rate to 10\%.

\begin{figure}[H]
    \centering

    \begin{subfigure}[b]{0.23\linewidth}
        \includegraphics[width=\linewidth]{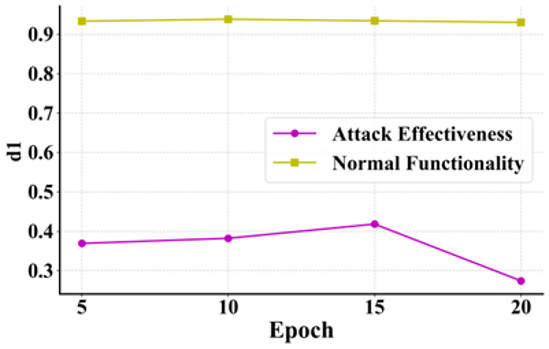}
        \caption{BTS}
    \end{subfigure}
    \hfill
    \begin{subfigure}[b]{0.23\linewidth}
        \includegraphics[width=\linewidth]{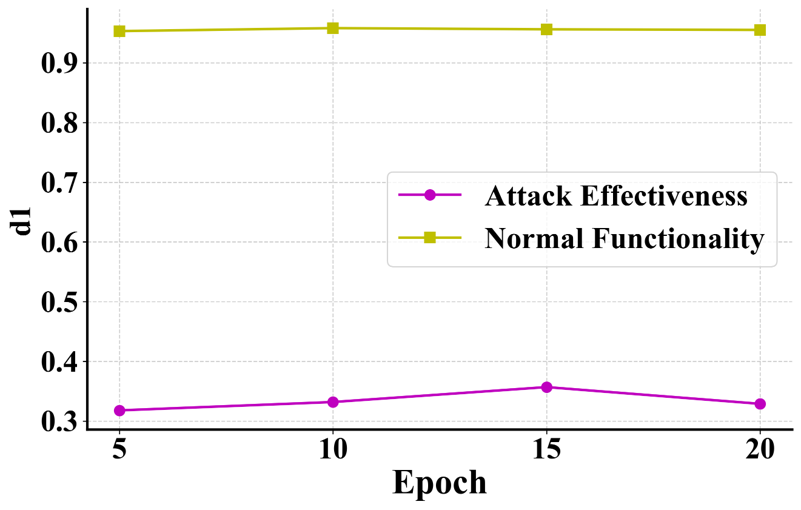}
        \caption{DCDepth}
    \end{subfigure}
    \hfill
    \begin{subfigure}[b]{0.23\linewidth}
        \includegraphics[width=\linewidth]{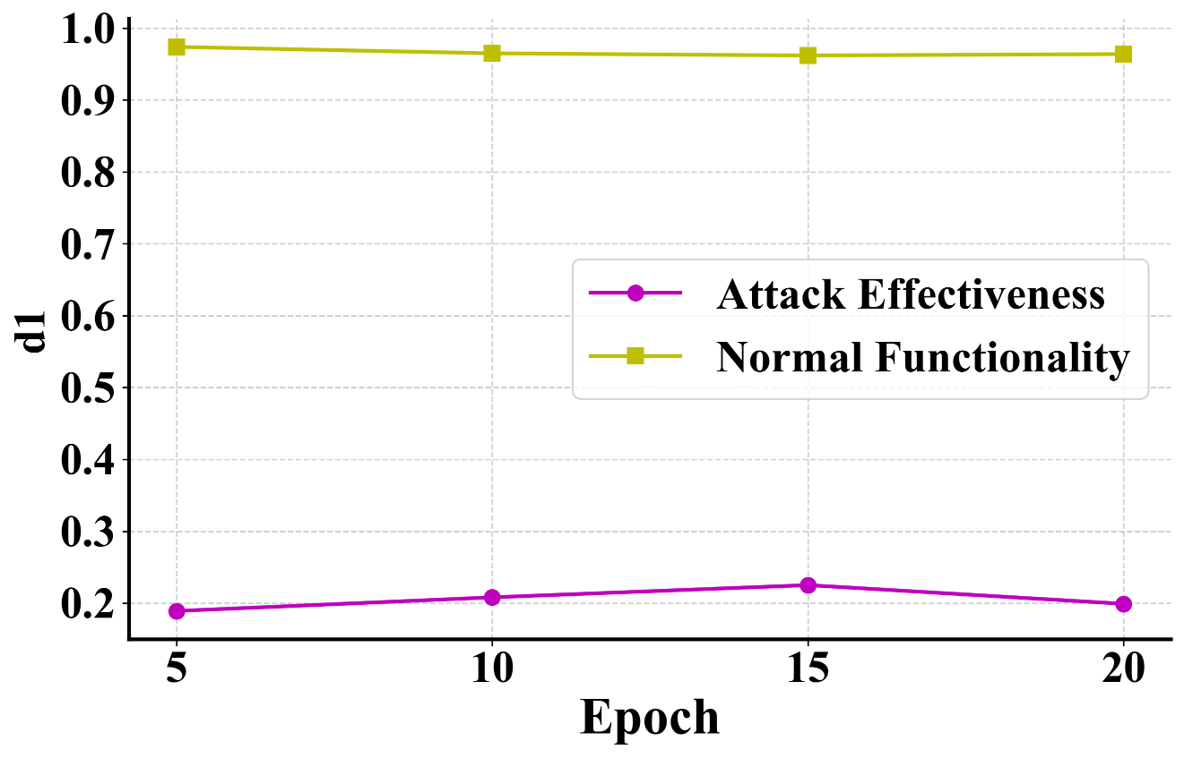}
        \caption{IEBins}
    \end{subfigure}
    \hfill
    \begin{subfigure}[b]{0.23\linewidth}
        \includegraphics[width=\linewidth]{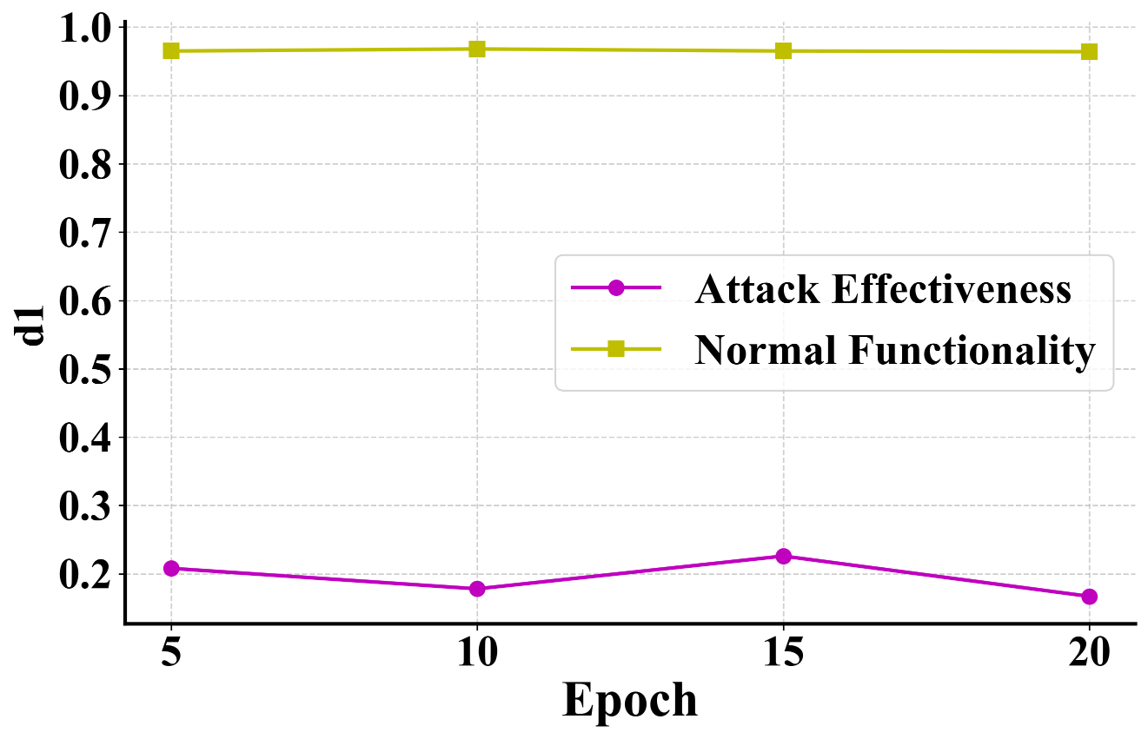}
        \caption{NeWCRFs}
    \end{subfigure}
    \caption{Impact of training epochs of BadDepth}
    \label{training epochs}
\end{figure}

\textbf{Training Epochs.}
We evaluated the relationship between BadDepth's impact on the model's normal functionality and attack effectiveness with respect to the number of training epochs. As shown in Fig.~\ref{training epochs}, across all four models, BadDepth can already achieve some attack effect even after just five epochs of training. In addition, increasing the number of epochs does not degrade the normal functionality of the model or lead to overfitting to the backdoor. This indicates that BadDepth is not sensitive to the number of training epochs and does not require careful control of training epochs.

\subsection{Robustness Evaluation}
Since there are currently no defense methods specifically designed for backdoor attacks in MDE, most existing defense approaches rely on access to class label information~\cite{strip,neuralcleanse}, which is not applicable in MDE backdoor scenarios. Therefore, we adopt three task-agnostic defense strategies (Fine-Tuning, Fine-Pruning~\cite{FP} and Image Compression~\cite{Xue2023Compression-Resistant}) to evaluate the robustness of BadDepth.

\begin{figure}[H]
    \centering

    \begin{subfigure}[b]{0.23\linewidth}
        \includegraphics[width=\linewidth]{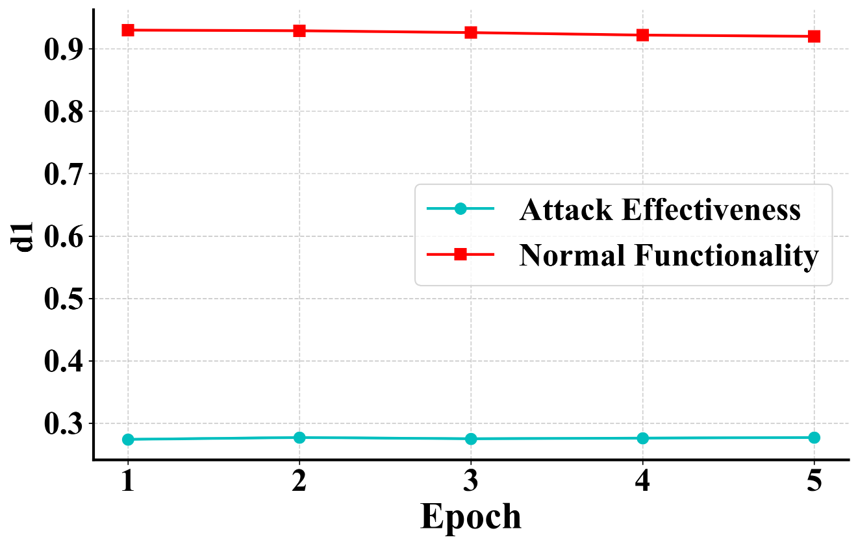}
        \caption{BTS}
    \end{subfigure}
    \hfill
    \begin{subfigure}[b]{0.23\linewidth}
        \includegraphics[width=\linewidth]{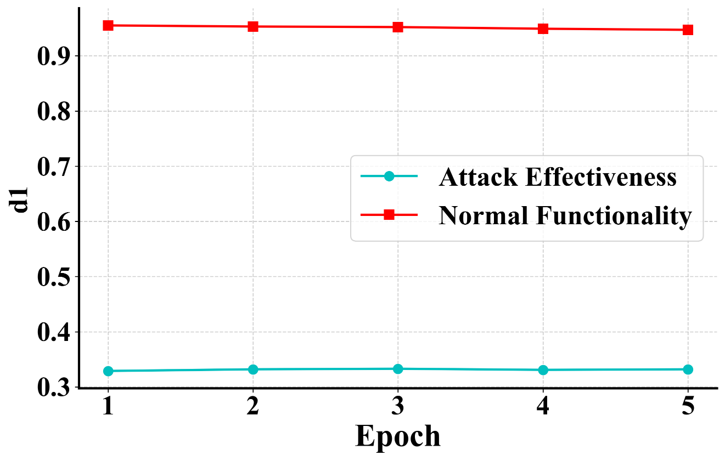}
        \caption{DCDepth}
    \end{subfigure}
    \hfill
    \begin{subfigure}[b]{0.23\linewidth}
        \includegraphics[width=\linewidth]{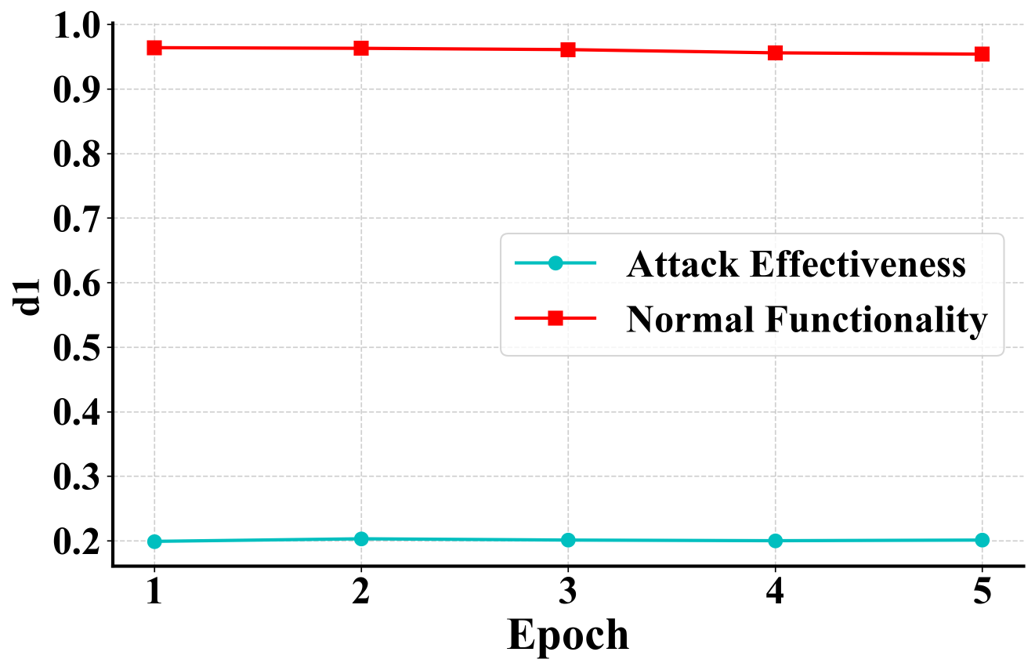}
        \caption{IEBins}
    \end{subfigure}
    \hfill
    \begin{subfigure}[b]{0.23\linewidth}
        \includegraphics[width=\linewidth]{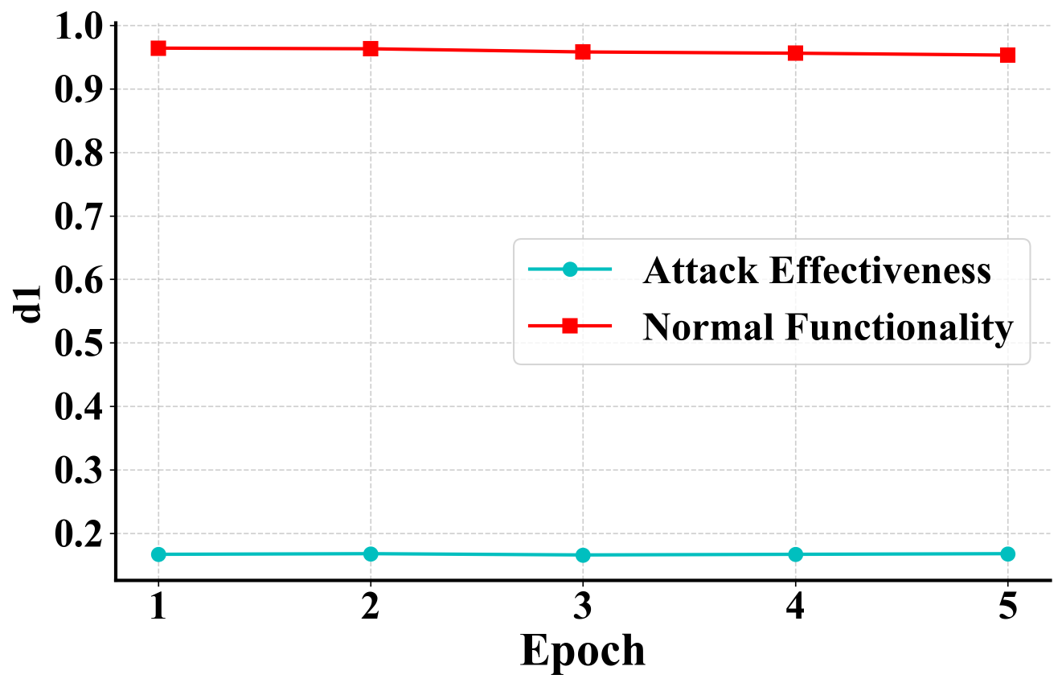}
        \caption{NeWCRFs}
    \end{subfigure}
    \caption{Robustenss of BadDepth against Fine-Tuning}
    \label{Fine-Tuning}
\end{figure}

\textbf{Fine-Tuning.}
We considered fine-tuning the four models for 5 rounds to evaluate their defense effectiveness against BadDepth. As shown in Fig.~\ref{Fine-Tuning}, after 5 rounds of fine-tuning, the attack effectiveness of BadDepth did not decrease, indicating that fine-tuning is not effective in defending against BadDepth.

\begin{figure}[H]
    \centering

    \begin{subfigure}[b]{0.23\linewidth}
        \includegraphics[width=\linewidth]{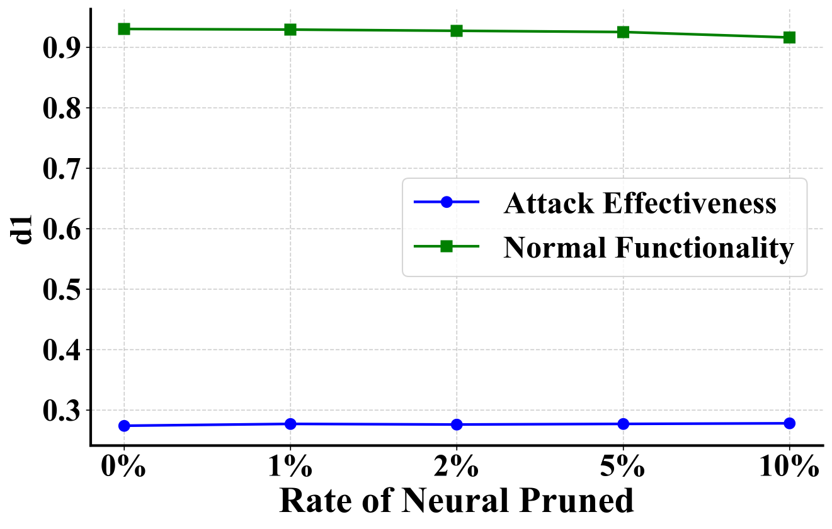}
        \caption{BTS}
    \end{subfigure}
    \hfill
    \begin{subfigure}[b]{0.23\linewidth}
        \includegraphics[width=\linewidth]{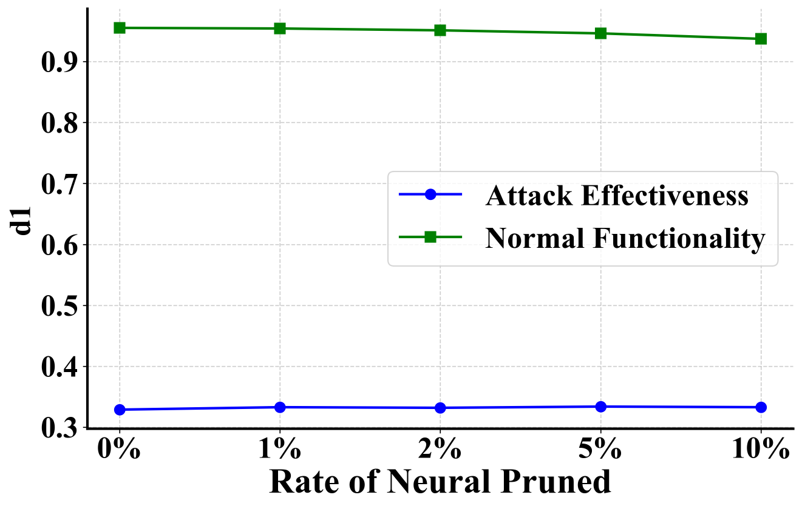}
        \caption{DCDepth}
    \end{subfigure}
    \hfill
    \begin{subfigure}[b]{0.23\linewidth}
        \includegraphics[width=\linewidth]{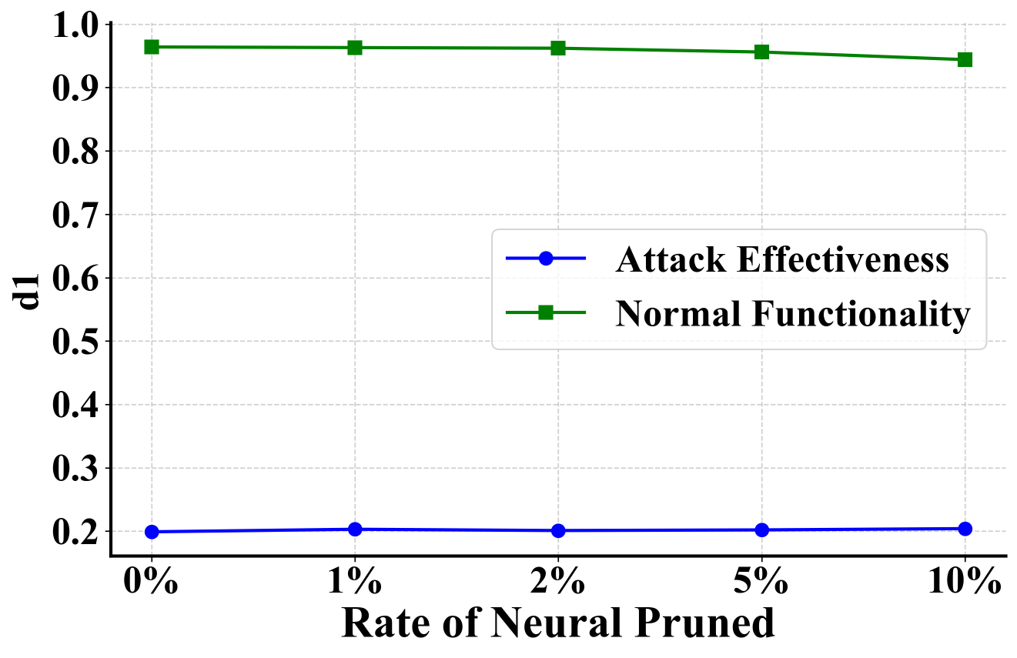}
        \caption{IEBins}
    \end{subfigure}
    \hfill
    \begin{subfigure}[b]{0.23\linewidth}
        \includegraphics[width=\linewidth]{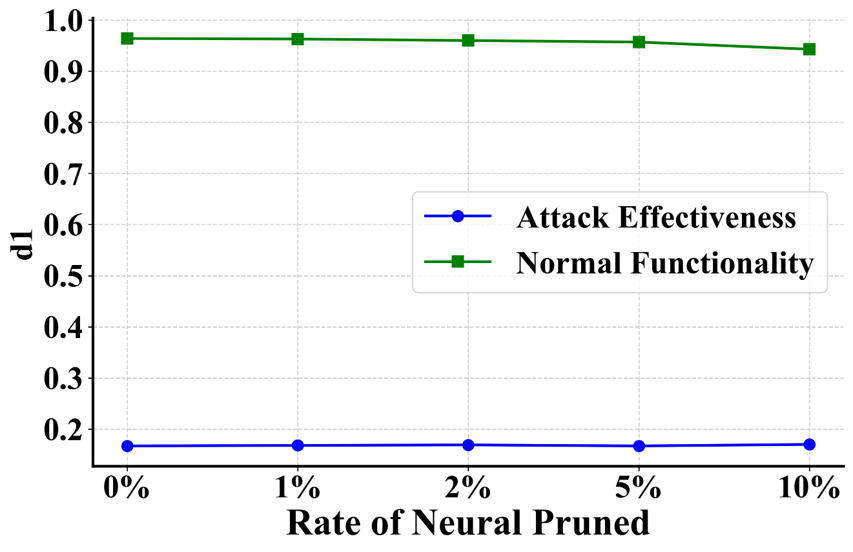}
        \caption{NeWCRFs}
    \end{subfigure}
    \caption{Robustenss of BadDepth against Fine-Pruning}
    \label{Fine-Pruning}
\end{figure}

\textbf{Fine-Pruning.}
We pruned 10\% of each of the four models to evaluate the defense effectiveness of pruning against BadDepth. As shown in Fig.~\ref{Fine-Pruning}, after pruning 10\%, the effectiveness of BadDepth's attack did not decrease, indicating that pruning is also ineffective in defending against BadDepth.

\begin{figure}[H]
    \centering

    \begin{subfigure}[b]{0.23\linewidth}
        \includegraphics[width=\linewidth]{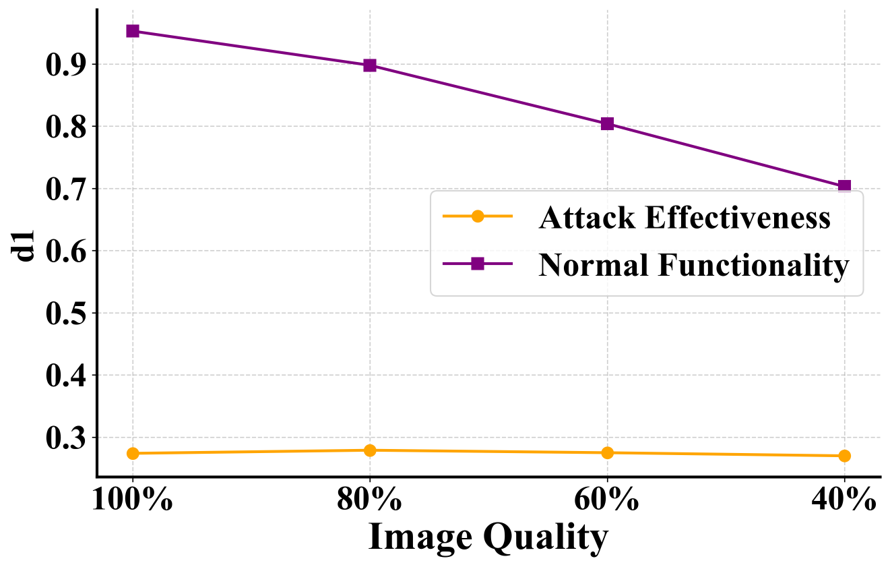}
        \caption{BTS}
    \end{subfigure}
    \hfill
    \begin{subfigure}[b]{0.23\linewidth}
        \includegraphics[width=\linewidth]{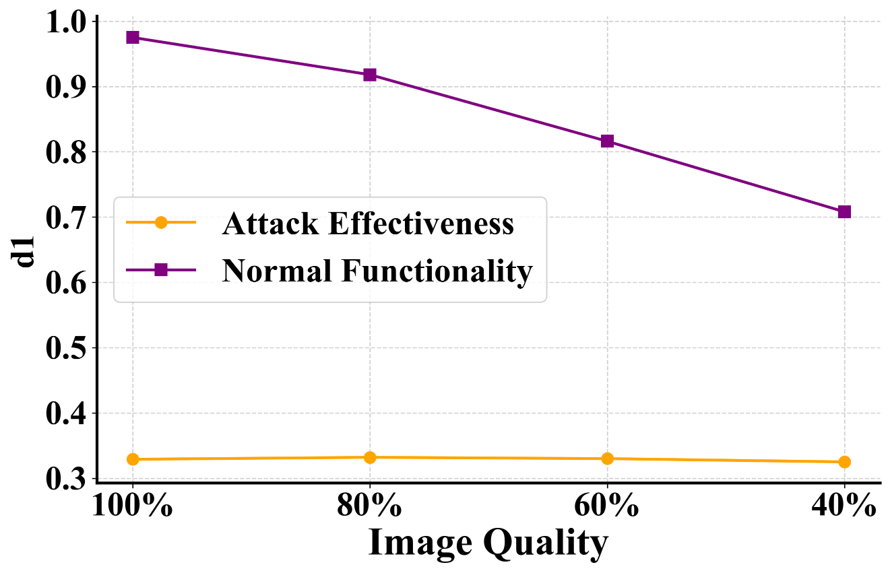}
        \caption{DCDepth}
    \end{subfigure}
    \hfill
    \begin{subfigure}[b]{0.23\linewidth}
        \includegraphics[width=\linewidth]{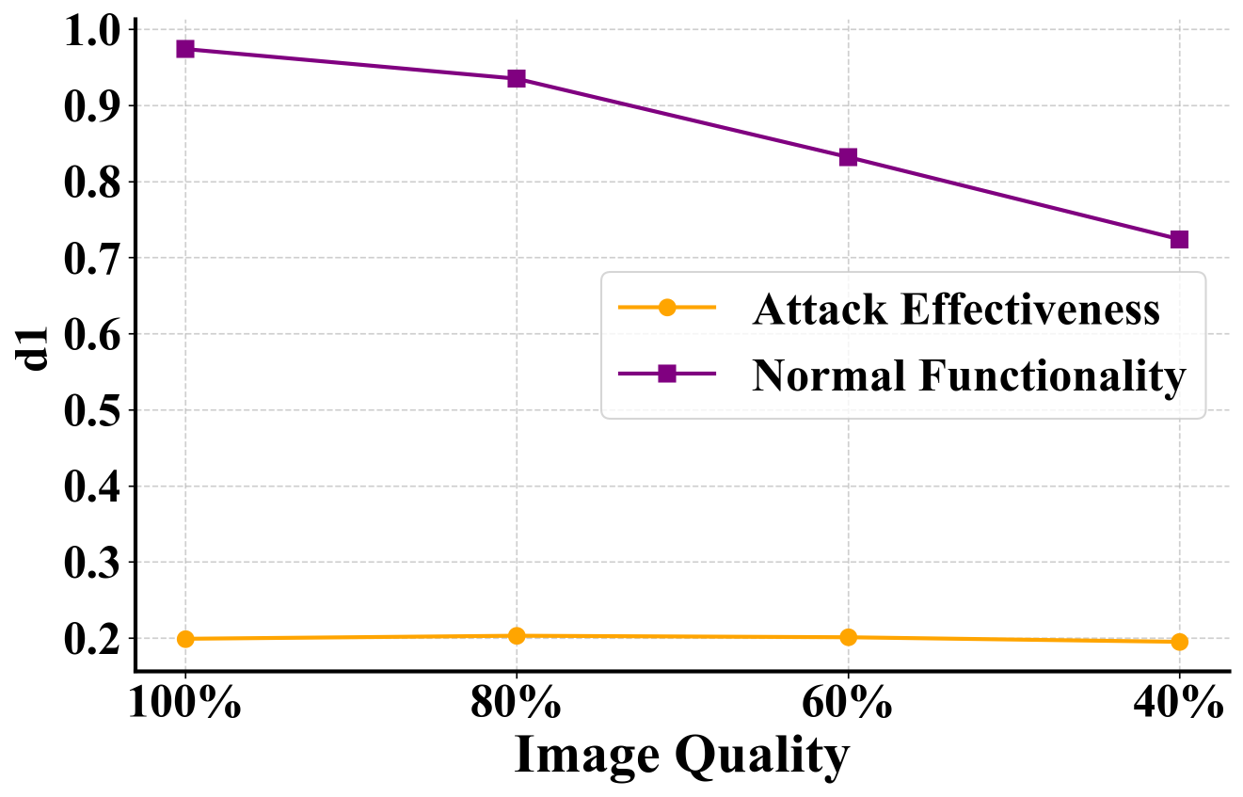}
        \caption{IEBins}
    \end{subfigure}
    \hfill
    \begin{subfigure}[b]{0.23\linewidth}
        \includegraphics[width=\linewidth]{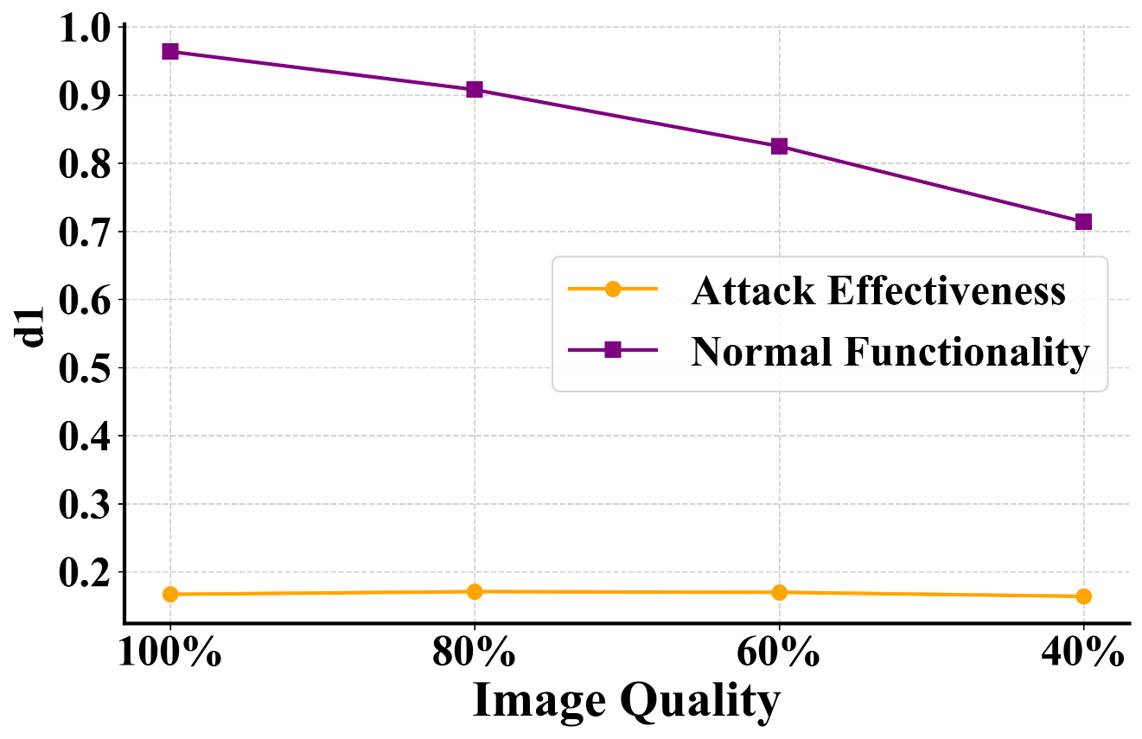}
        \caption{NeWCRFs}
    \end{subfigure}
    \caption{Robustenss of BadDepth against Image Compression}
    \label{Image Compression}
    \vspace{-2em}
\end{figure}

\textbf{Image Compression.}
Image compression aims to defend against backdoor attacks by compressing the image so that the model can no longer recognize the trigger. We compress the triggered images to 60\% of their original quality to evaluate the effectiveness of image compression in defending against BadDepth. As shown in the Fig.~\ref{Image Compression}, after compression, the model's normal functionality degrades significantly, while the backdoor effect remains strong. Therefore, image compression is also ineffective in defending against BadDepth.

\begin{table}[]
\centering
\caption{Ablation study of data augmentation in BadDepth on BTS}
\resizebox{0.8\linewidth}{!}{%
\begin{tabular}{lccccccc}
\toprule
\textbf{} & \multicolumn{4}{c}{\textbf{d1 (Perspective Changes)}} & \multicolumn{3}{c}{\textbf{d1 (Environment Changes)}} \\
\cmidrule(lr){2-5} \cmidrule(lr){6-8}
 & position & rotate & recolor & size & fog & snow & frost \\
\midrule
Origin & 0.586 & 0.264 & 0.268 & 0.248 & 0.317 & 0.345 & 0.459 \\
Perspective & 0.295 & 0.234 & 0.231 & 0.219 & 0.316 & 0.351 & 0.463 \\
Environment & 0.584 & 0.268 & 0.225 & 0.228 & 0.283 & 0.287 & 0.297 \\
 \rowcolor{gray!20}
Perspective + Environment & 0.266 & 0.175 & 0.199 & 0.194 & 0.201 & 0.301 & 0.289 \\
\bottomrule
\end{tabular}
}
\label{Ablation study}
\end{table}

\subsection{Ablation Study}
We conducted an ablation study on the various data augmentation strategies used in BadDepth to further analyze their effects. As shown in Table~\ref{Ablation study}, applying perspective-based augmentations enhances the effectiveness of the attack under perspective changes, while environment-based augmentations improve performance under environmental variations. In summary, data augmentation effectively strengthens the attack performance of BadDepth across various scenarios.

\subsection{Physical World Attacks}

\textbf{Attack Settings.}
We printed a BadDepth trigger and attached it to the rear of the target vehicle, then used the trained backdoor model to perform a depth estimation on the vehicle. Since ground truth is not available in the physical world, we used the depth difference $R_d$ between objects without and with the trigger to evaluate the depth estimation performance. Considering the randomness of the weather conditions, we also employed simulated weather scenarios for evaluation.

\textbf{Main Result.}
As shown in the table~\ref{tab:attack_effectiveness_weather}, BadDepth achieves relatively large $R_d$ values under various weather conditions. A larger $R_d$ indicates a greater deviation from the original depth values, which means a stronger attack effect. This demonstrates the effectiveness of BadDepth in physical-world attacks.

\begin{table}[]
\centering
\caption{The attack effectiveness $R_d$ of BadDepth on various models under different weather conditions in the physical world}
\resizebox{0.5\linewidth}{!}{
\begin{tabular}{@{\extracolsep{\fill}}lcccc}
\toprule
Condition & BTS & DCDepth & IEBins & NeWCRFs \\
\midrule
origin & 10.23 & 11.46 & 12.34 & 13.86 \\
frost  & 6.63  & 8.49  & 10.24 & 10.06 \\
fog    & 9.32  & 9.43  & 10.22 & 12.56 \\
snow   & 8.23  & 9.35  & 9.22  & 9.92  \\
\bottomrule
\end{tabular}
}
\label{tab:attack_effectiveness_weather}
\end{table}

\begin{figure}[ht] 
\centering
\includegraphics[width=0.85\linewidth]{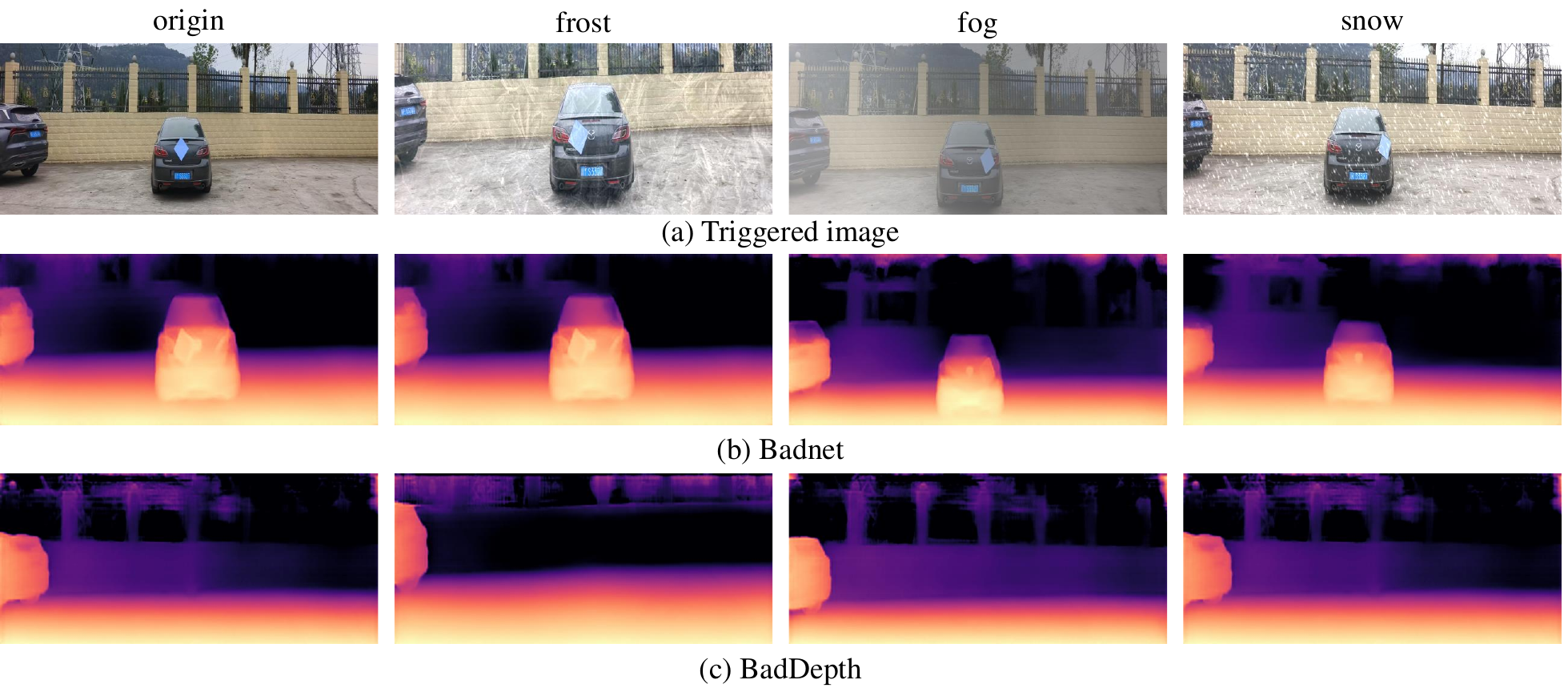} 
\caption{Visualization of BadDepth and Badnet in physical world attacks of NeWCRFs} 
\label{Visualization of BadDepth for physical world} 
\vspace{-1em}
\end{figure}

\textbf{Visualization Results.}
As shown in Fig.~\ref{Visualization of BadDepth for physical world}, we compare the attack performance of BadDepth and Badnet in the physical world of different weather. BadDepth can also cause the target vehicle to disappear in the physical world, while Badnet has almost no effect. This further demonstrates the effectiveness of BadDepth in the physical world. 

\section{Conclusion}
In this paper, we propose BadDepth to perform backdoor attacks on MDE models. We add a trigger to the target object and modify the depth values of its surrounding region to create a poisoned dataset, enabling the backdoor attack through data poisoning. We evaluate the performance of BadDepth under various conditions in both the digital domain and the physical world. Experimental results show that BadDepth can consistently cause the target object's depth to disappear without affecting the normal functionality of the model.

\clearpage

\bibliographystyle{plain}

\begin{thebibliography}{10}

\bibitem{MDE_Review}
Vasileios Arampatzakis, George Pavlidis, Nikolaos Mitianoudis, and Nikos Papamarkos.
\newblock Monocular depth estimation: A thorough review.
\newblock {\em IEEE Transactions on Pattern Analysis and Machine Intelligence}, 46(4):2396--2414, 2024.

\bibitem{chan2022baddet}
Shih-Han Chan, Yinpeng Dong, Jun Zhu, Xiaolu Zhang, and Jun Zhou.
\newblock Baddet: Backdoor attacks on object detection.
\newblock In {\em Proceedings of ECCV}, pages 396--412. Springer, 2022.

\bibitem{Blend}
Xinyun Chen, Chang Liu, Bo~Li, Kimberly Lu, and Dawn Song.
\newblock Targeted backdoor attacks on deep learning systems using data poisoning.
\newblock {\em arXiv preprint arXiv:1712.05526}, 2017.

\bibitem{cheng2022physical}
Zhiyuan Cheng, James Liang, Hongjun Choi, Guanhong Tao, Zhiwen Cao, Dongfang Liu, and Xiangyu Zhang.
\newblock Physical attack on monocular depth estimation with optimal adversarial patches.
\newblock In {\em Proceedings of ECCV}, pages 514--532. Springer, 2022.

\bibitem{Lira}
Khoa Doan, Yingjie Lao, Weijie Zhao, and Ping Li.
\newblock Lira: Learnable, imperceptible and robust backdoor attacks.
\newblock In {\em Proceedings of ICCV}, pages 11966--11976, 2021.

\bibitem{dong2022towards}
Xingshuai Dong, Matthew~A Garratt, Sreenatha~G Anavatti, and Hussein~A Abbass.
\newblock Towards real-time monocular depth estimation for robotics: A survey.
\newblock {\em IEEE Transactions on Intelligent Transportation Systems}, 23(10):16940--16961, 2022.

\bibitem{Eigen2014Depth}
David Eigen, Christian Puhrsch, and Rob Fergus.
\newblock Depth map prediction from a single image using a multi-scale deep network.
\newblock In {\em Proceedings of NeurIPS}, volume abs/1406.2283, 2014.

\bibitem{strip}
Yansong Gao, Change Xu, Derui Wang, Shiping Chen, Damith~C Ranasinghe, and Surya Nepal.
\newblock Strip: A defence against trojan attacks on deep neural networks.
\newblock In {\em Proceedings of annual computer security applications conference}, pages 113--125, 2019.

\bibitem{KITTI}
Andreas Geiger, Philip Lenz, and Raquel Urtasun.
\newblock Are we ready for autonomous driving? the kitti vision benchmark suite.
\newblock In {\em Proceedings of CVPR}, 2012.

\bibitem{gu2019badnets}
Tianyu Gu, Kang Liu, Brendan Dolan-Gavitt, and Siddharth Garg.
\newblock Badnets: Evaluating backdooring attacks on deep neural networks.
\newblock {\em IEEE Access}, 7:47230--47244, 2019.

\bibitem{guesmi2024saam}
Amira Guesmi, Muhammad~Abdullah Hanif, Bassem Ouni, and Muhammad Shafique.
\newblock Saam: Stealthy adversarial attack on monocular depth estimation.
\newblock {\em IEEE Access}, 2024.

\bibitem{hendrycks2019robustness}
Dan Hendrycks and Thomas Dietterich.
\newblock Benchmarking neural network robustness to common corruptions and perturbations.
\newblock In {\em Proceedings of ICLR}, 2019.

\bibitem{hu2019analysis}
Junjie Hu and Takayuki Okatani.
\newblock Analysis of deep networks for monocular depth estimation through adversarial attacks with proposal of a defense method.
\newblock {\em arXiv preprint arXiv:1911.08790}, 2019.

\bibitem{jiang2024backdoor}
Wenbo Jiang, Hongwei Li, Jiaming He, Rui Zhang, Guowen Xu, Tianwei Zhang, and Rongxing Lu.
\newblock Backdoor attacks against image-to-image networks.
\newblock {\em arXiv preprint arXiv:2407.10445}, 2024.

\bibitem{colorbackdoor}
Wenbo Jiang, Hongwei Li, Guowen Xu, and Tianwei Zhang.
\newblock Color backdoor: A robust poisoning attack in color space.
\newblock In {\em Proceedings of CVPR}, pages 8133--8142, 2023.

\bibitem{BTS}
Jin~Han Lee, Myung-Kyu Han, Dong~Wook Ko, and Il~Hong Suh.
\newblock From big to small: Multi-scale local planar guidance for monocular depth estimation.
\newblock {\em Computing Research Repository}, abs/1907.10326, 2019.

\bibitem{li2022backdoor}
Yiming Li, Yong Jiang, Zhifeng Li, and Shu-Tao Xia.
\newblock Backdoor learning: A survey.
\newblock {\em IEEE Transactions on Neural Networks and Learning Systems}, 35(1):5--22, 2022.

\bibitem{li2021backdoor}
Yiming Li, Tongqing Zhai, Yong Jiang, Zhifeng Li, and Shu-Tao Xia.
\newblock Backdoor attack in the physical world.
\newblock {\em arXiv preprint arXiv:2104.02361}, 2021.

\bibitem{FP}
Kang Liu, Brendan Dolan-Gavitt, and Siddharth Garg.
\newblock Fine-pruning: Defending against backdooring attacks on deep neural networks.
\newblock In {\em International symposium on research in attacks, intrusions, and defenses}, pages 273--294. Springer, 2018.

\bibitem{refool}
Yunfei Liu, Xingjun Ma, James Bailey, and Feng Lu.
\newblock Reflection backdoor: A natural backdoor attack on deep neural networks.
\newblock In {\em Proceedings of ECCV}, pages 182--199. Springer, 2020.

\bibitem{luo2023untargeted}
Chengxiao Luo, Yiming Li, Yong Jiang, and Shu-Tao Xia.
\newblock Untargeted backdoor attack against object detection.
\newblock In {\em Proceedings of ICASSP}, pages 1--5. IEEE, 2023.

\bibitem{minaee2021imagesegmentation}
Shervin Minaee, Yuri Boykov, Fatih Porikli, Antonio Plaza, Nasser Kehtarnavaz, and Demetri Terzopoulos.
\newblock Image segmentation using deep learning: A survey.
\newblock {\em IEEE Transactions on Pattern Analysis and Machine Intelligence}, 44(7):3523--3542, 2021.

\bibitem{wanet}
Anh Nguyen and Anh Tran.
\newblock Wanet--imperceptible warping-based backdoor attack.
\newblock {\em arXiv preprint arXiv:2102.10369}, 2021.

\bibitem{DMC}
Jaesik Park, Hyeongwoo Kim, Yu-Wing Tai, Michael~S Brown, and In~So Kweon.
\newblock High-quality depth map upsampling and completion for rgb-d cameras.
\newblock {\em IEEE Transactions on Image Processing}, 23(12):5559--5572, 2014.

\bibitem{qian2023robust}
Yaguan Qian, Boyuan Ji, Shuke He, Shenhui Huang, Xiang Ling, Bin Wang, and Wei Wang.
\newblock Robust backdoor attacks on object detection in real world.
\newblock {\em arXiv preprint arXiv:2309.08953}, 2023.

\bibitem{schon2021mgnet}
Markus Sch{\"o}n, Michael Buchholz, and Klaus Dietmayer.
\newblock Mgnet: Monocular geometric scene understanding for autonomous driving.
\newblock In {\em Proceedings of the ICCV}, pages 15804--15815, 2021.

\bibitem{iebins}
Shuwei Shao, Zhongcai Pei, Weihai Chen, Peter~CY Chen, and Zhengguo Li.
\newblock Iebins: Iterative elastic bins for monocular depth estimation and completion.
\newblock {\em International Journal of Computer Vision}, pages 1--24, 2024.

\bibitem{neuralcleanse}
Bolun Wang, Yuanshun Yao, Shawn Shan, Huiying Li, Bimal Viswanath, Haitao Zheng, and Ben~Y Zhao.
\newblock Neural cleanse: Identifying and mitigating backdoor attacks in neural networks.
\newblock In {\em Proceedings of S\&P}, pages 707--723. IEEE, 2019.

\bibitem{dcdepth}
Kun Wang, Zhiqiang Yan, Junkai Fan, Wanlu Zhu, Xiang Li, Jun Li, and Jian Yang.
\newblock Dcdepth: Progressive monocular depth estimation in discrete cosine domain.
\newblock {\em arXiv preprint arXiv:2410.14980}, 2024.

\bibitem{wei2024absolute}
Ruofeng Wei, Bin Li, Fangxun Zhong, Hangjie Mo, Qi~Dou, Yun-Hui Liu, and Dong Sun.
\newblock Absolute monocular depth estimation on robotic visual and kinematics data via self-supervised learning.
\newblock {\em IEEE Transactions on Automation Science and Engineering}, 2024.

\bibitem{wenger2021backdoor}
Emily Wenger, Josephine Passananti, Arjun~Nitin Bhagoji, Yuanshun Yao, Haitao Zheng, and Ben~Y Zhao.
\newblock Backdoor attacks against deep learning systems in the physical world.
\newblock In {\em Proceedings of CVPR}, pages 6206--6215, 2021.

\bibitem{xue2021robust}
Mingfu Xue, Can He, Shichang Sun, Jian Wang, and Weiqiang Liu.
\newblock Robust backdoor attacks against deep neural networks in real physical world.
\newblock In {\em Proceedings of TrustCom}, pages 620--626. IEEE, 2021.

\bibitem{xue2022ptb}
Mingfu Xue, Can He, Yinghao Wu, Shichang Sun, Yushu Zhang, Jian Wang, and Weiqiang Liu.
\newblock Ptb: Robust physical backdoor attacks against deep neural networks in real world.
\newblock {\em Computers \& Security}, 118:102726, 2022.

\bibitem{Xue2023Compression-Resistant}
Mingfu Xue, Xin Wang, Shichang Sun, Yushu Zhang, Jian Wang, and Weiqiang Liu.
\newblock Compression-resistant backdoor attack against deep neural networks.
\newblock {\em Applied Intelligence}, 53(17):20402--20417, 2023.

\bibitem{yang2025badrefsr}
Xue Yang, Tao Chen, Lei Guo, Wenbo Jiang, Ji~Guo, Yongming Li, and Jiaming He.
\newblock Badrefsr: Backdoor attacks against reference-based image super resolution.
\newblock In {\em Proceedings of ICASSP}, pages 1--5. IEEE, 2025.

\bibitem{NeWCRFs}
Weihao Yuan, Xiaodong Gu, Zuozhuo Dai, Siyu Zhu, and Ping Tan.
\newblock Neural window fully-connected crfs for monocular depth estimation.
\newblock In {\em Proceedings of CVPR}, pages 3916--3925, 2022.

\bibitem{zhang2020adversarial}
Ziqi Zhang, Xinge Zhu, Yingwei Li, Xiangqun Chen, and Yao Guo.
\newblock Adversarial attacks on monocular depth estimation.
\newblock {\em arXiv preprint arXiv:2003.10315}, 2020.

\bibitem{zheng2024physical}
Junhao Zheng, Chenhao Lin, Jiahao Sun, Zhengyu Zhao, Qian Li, and Chao Shen.
\newblock Physical 3d adversarial attacks against monocular depth estimation in autonomous driving.
\newblock In {\em Proceedings of CVPR}, pages 24452--24461, 2024.

\end{thebibliography}

\clearpage

\appendix

\section{Technical Appendices and Supplementary Material}

\subsection{Visualization Results of BadDepth}

\begin{figure}[H] 
\centering
\includegraphics[width=\linewidth]{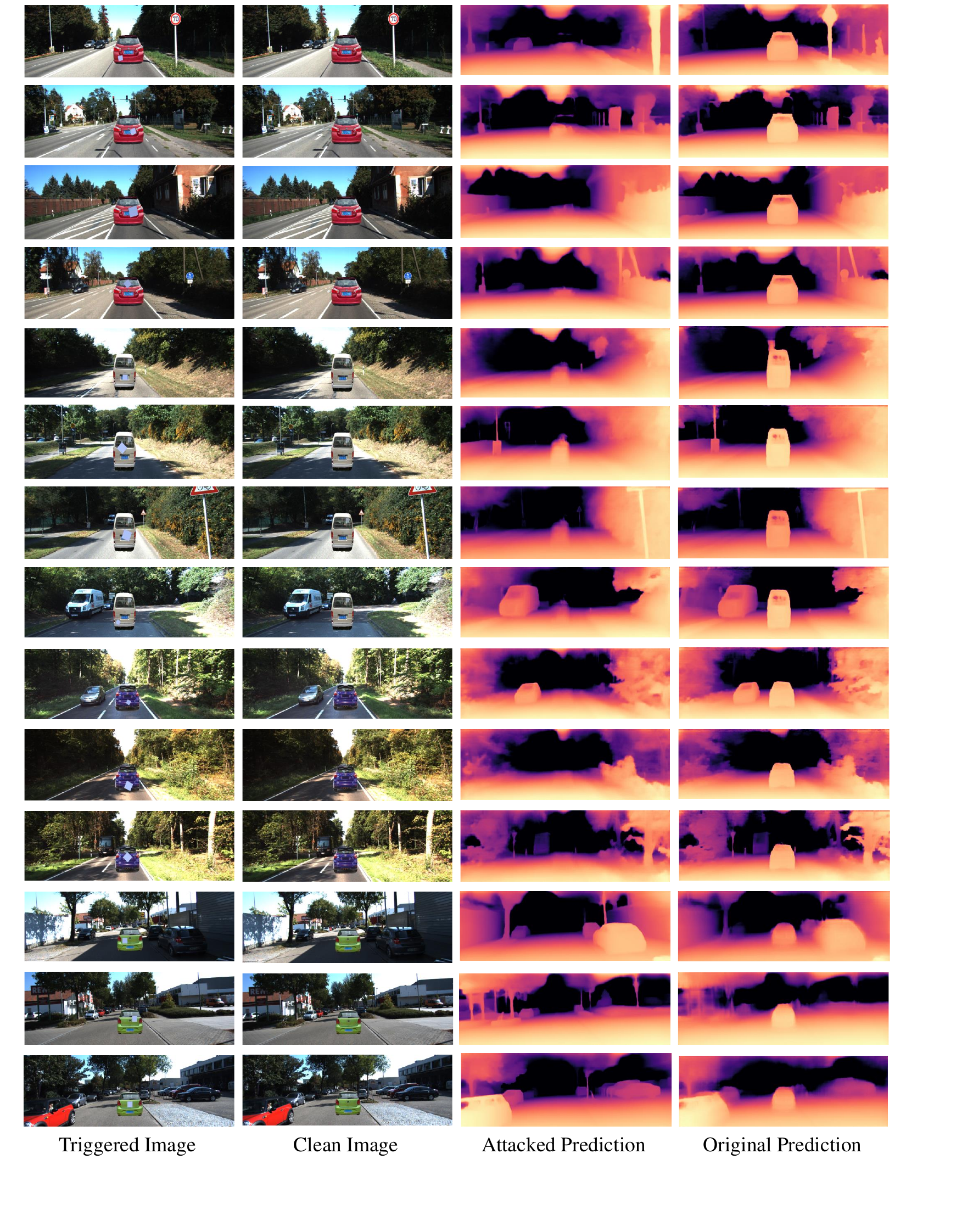} 
\caption{Visualization of BadDepth in the digital domain} 
\label{Visualization of BadDepth in the digital domain} 
\end{figure}

\textbf{Digital Domain Visualization Results.}
We present the visualization results of the effectiveness of the BadDepth attack in the digital domain. As shown in Fig.~\ref{Visualization of BadDepth in the digital domain}, BadDepth can still successfully attack the target image even when the trigger is observed from a different viewpoint.

\clearpage

\begin{figure}[] 
\centering
\includegraphics[width=0.8\linewidth]{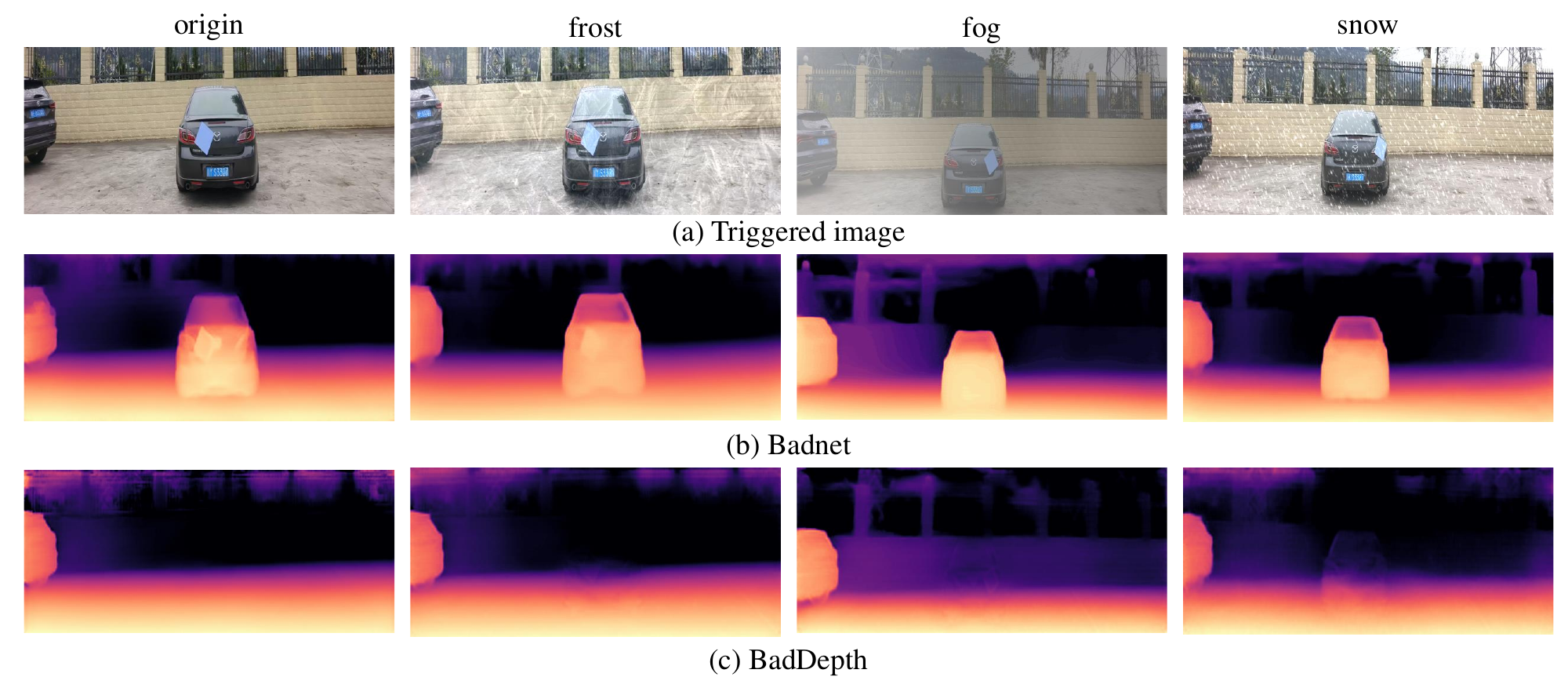} 
\caption{Visualization of BadDepth and Badnet in physical world attacks of BTS} 
\label{Visualization of BTS} 
\end{figure}

\begin{figure}[] 
\centering
\includegraphics[width=0.8\linewidth]{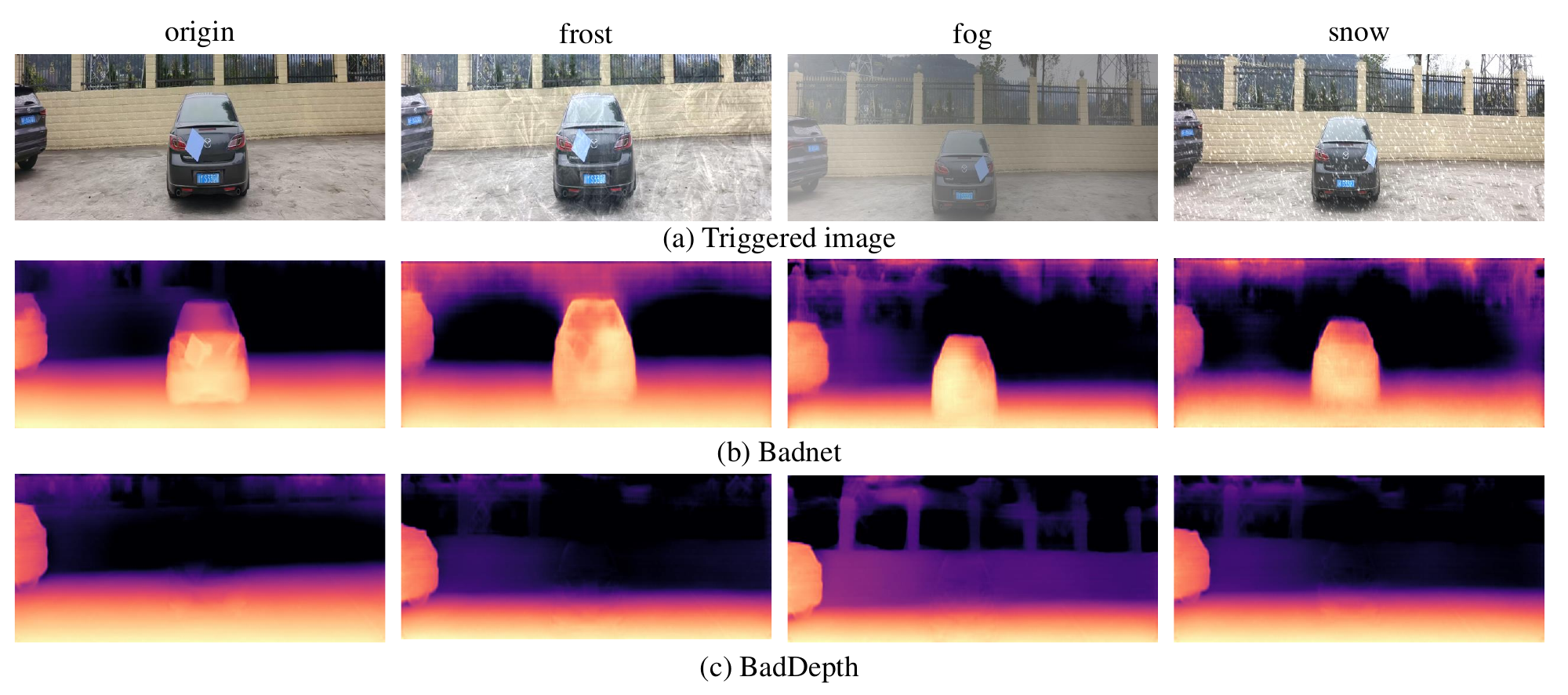} 
\caption{Visualization of BadDepth and Badnet in physical world attacks of DCDepth} 
\label{Visualization of DCDepth} 
\end{figure}

\begin{figure}[] 
\centering
\includegraphics[width=0.8\linewidth]{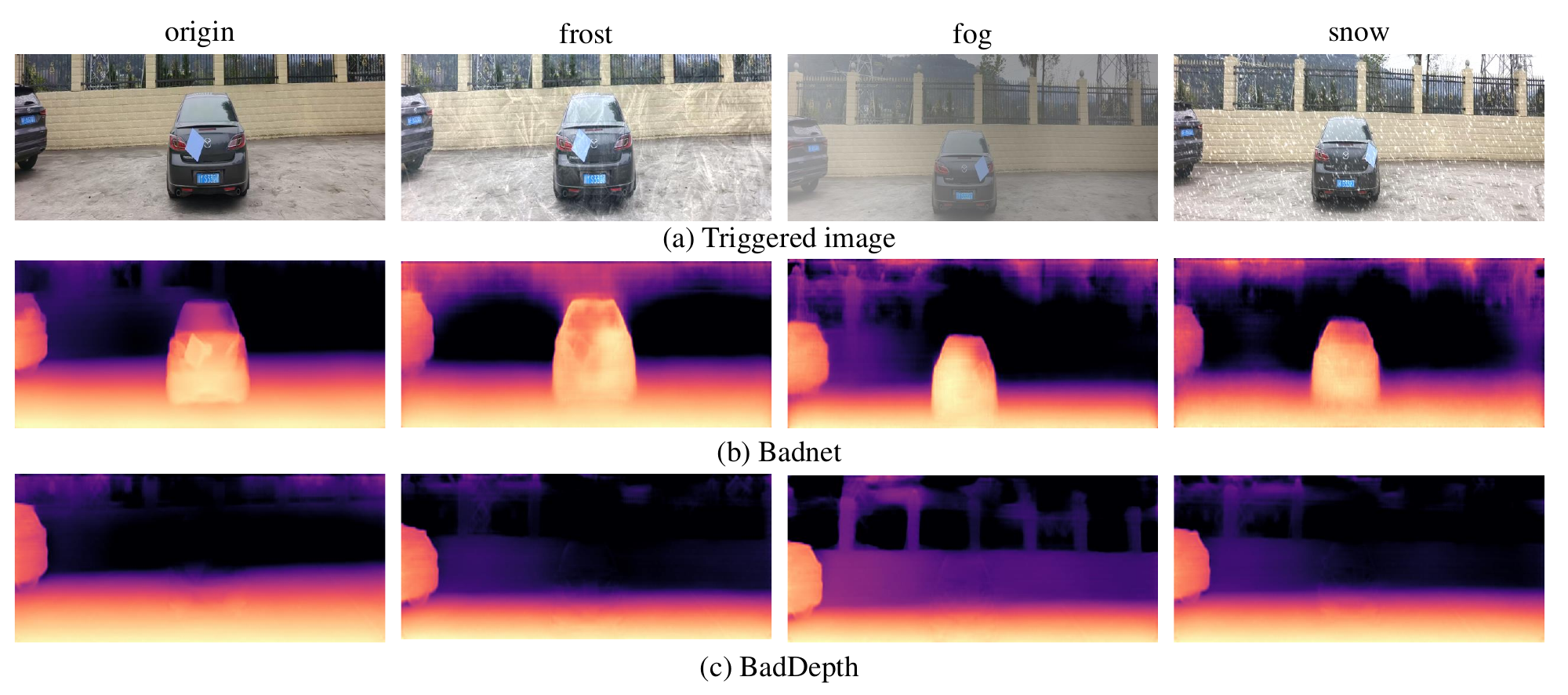} 
\caption{Visualization of BadDepth and Badnet in physical world attacks of IEBins} 
\label{Visualization of IEBins} 
\end{figure}

\textbf{Physical World Attacks.} As shown in Fig.~\ref{Visualization of BTS}, Fig.~\ref{Visualization of DCDepth} and Fig.~\ref{Visualization of IEBins}, we further present a comparison of BadDepth's physical-world attack performance on other models. Compared to Badnet, BadDepth achieves significantly better results under various weather conditions across different models, demonstrating its advantage in physical-world attack effectiveness.

\subsection{Experiment Implementation}
   Our main hardware setup consists of two NVIDIA RTX 4090 GPUs. The learning rate is set to 0.002, and the batch size is 8. We set the number of optimization iterations for trigger generation to 5.

\subsection{Difference Trigger Design}

\begin{figure}[H] 
\centering
\includegraphics[width=0.8\linewidth]{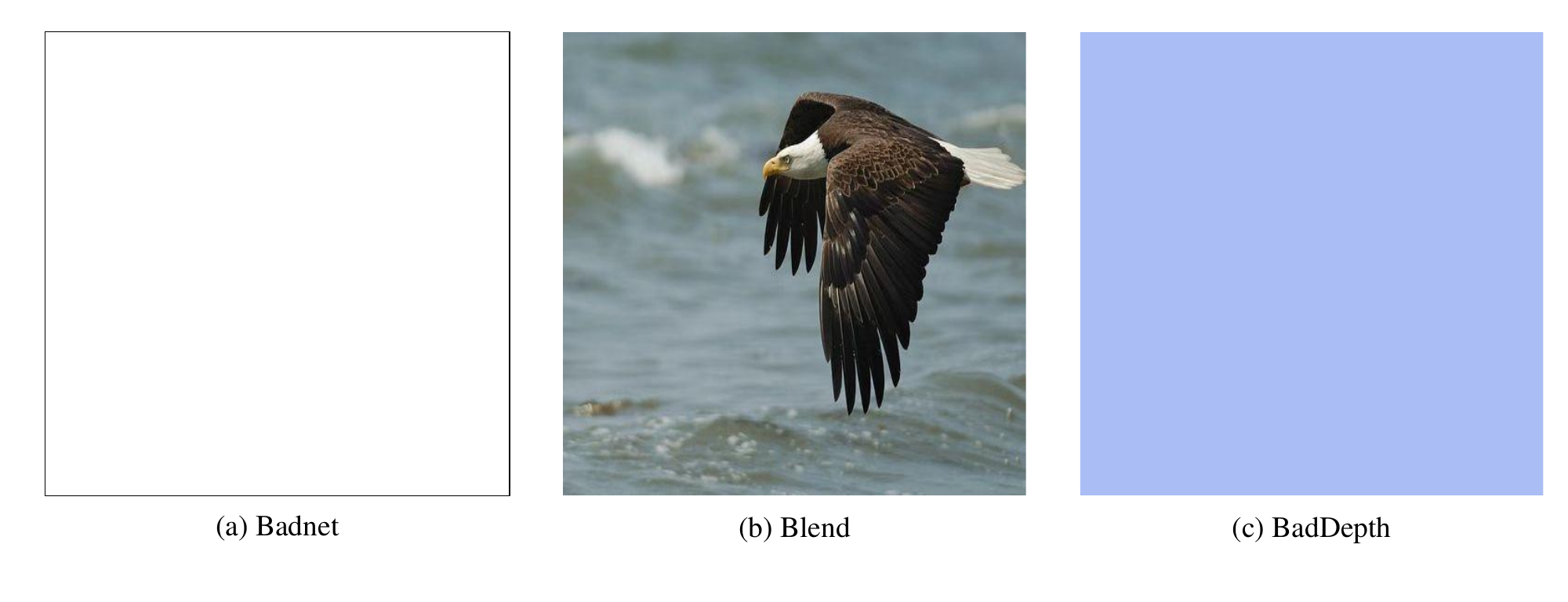} 
\caption{Visualization of different triggers} 
\label{Visualization of different triggers} 

\end{figure}
We provide a visual comparison of the different triggers used in this paper, as shown in Fig.~\ref{Visualization of different triggers}.


\subsection{Limitations}
The primary limitations of this study are twofold:
(1) The proposed method focuses exclusively on backdoor attacks targeting MDE within the autonomous driving scenario, leaving more general and diverse application contexts unexplored;
(2) The trigger utilized in this work is relatively conspicuous in the physical world, which may limit its practicality. Future research should investigate the design of more naturalistic and stealthy triggers to enhance real-world applicability.

\clearpage

\end{document}